\ifdefined\arxivpreview\else
\documentclass{adobe_research}
\fi

\usepackage[T1]{fontenc}
\usepackage{lmodern}
\setcitestyle{authoryear,round,citesep={;},aysep={,},yysep={;}}

\usepackage{amsmath,amsfonts,bm}

\def\eqref#1{equation~\ref{#1}}

\def\1{\bm{1}}

\DeclareMathAlphabet{\mathsfit}{\encodingdefault}{\sfdefault}{m}{sl}
\SetMathAlphabet{\mathsfit}{bold}{\encodingdefault}{\sfdefault}{bx}{n}

\usepackage{array}
\usepackage{wrapfig}
\usepackage{url}
\usepackage{xurl}
\usepackage{listings}
\usepackage{needspace}
\usepackage{float}
\hypersetup{
  colorlinks=true,
  linkcolor=blue,
  citecolor=adobered,
  urlcolor=adobered!45!black,
  pdftitle={OmniXRef: A Unified Framework for Multimodal Multi-Reference Video Editing},
  pdfauthor={Xiaoyan Cong, Tianyu Wang, Hyeonho Jeong, Zhifei Zhang, Yuqian Zhou, Yijun Li, Shaoteng Liu, Zhe Lin, Srinath Sridhar, Soo Ye Kim}
}

\definecolor{promptframe}{gray}{0.45}
\definecolor{promptback}{rgb}{0.965,0.972,0.988}
\lstdefinestyle{promptjson}{basicstyle=\fontsize{6.8}{8.2}\selectfont\ttfamily\color{black!75},
  columns=fixed, keepspaces=true, breaklines=true, showstringspaces=false,
  aboveskip=2pt, belowskip=0pt, xleftmargin=0.8em}
\newcommand{\promptrole}[1]{\par\vspace{4pt}\noindent\textbf{#1}\par\vspace{1pt}}
\newcommand{\prompthead}[1]{\par\vspace{2pt}\noindent\textbf{#1.}~}
\newcommand{\promptattach}[1]{\par\vspace{2pt}\noindent #1\par}
\newcommand{\promptreturn}{\par\vspace{3pt}\noindent\textbf{Return only:}\par}
\newenvironment{promptitems}{\begin{list}{}{\setlength{\leftmargin}{1.2em}\setlength{\itemindent}{-0.6em}%
  \setlength{\labelwidth}{0pt}\setlength{\labelsep}{0.4em}\setlength{\topsep}{1pt}\setlength{\itemsep}{0.5pt}%
  \setlength{\parsep}{0pt}}}{\end{list}}

\newcommand{\frameworkname}{OmniXRef}
\newcommand{\modelname}{\frameworkname-Model}
\newcommand{\benchname}{\frameworkname-Bench}
\newcommand{\dataname}{\frameworkname-Data}
\newcommand{\rropername}{R\textsuperscript{2}-RoPE}
\ExplSyntaxOn
\NewDocumentCommand{\editsetting}{m}{\ensuremath{\tl_map_inline:nn {#1} {\mathrm{##1}'}\mathrm{V2V}}}
\ExplSyntaxOff
\def\editsettingpdfrefs#1{\ifx#1\relax\else#1'\expandafter\editsettingpdfrefs\fi}
\title{\frameworkname{}: A Unified Framework for Multimodal Multi-Reference Video Editing}
\author[1*]{Xiaoyan Cong}
\author[2]{Tianyu Wang}
\author[2]{Hyeonho Jeong}
\author[2]{Zhifei Zhang}
\author[2]{Yuqian Zhou}
\authorbreak
\author[2]{Yijun Li}
\author[2]{Shaoteng Liu}
\author[2]{Zhe Lin}
\author[1]{Srinath Sridhar}
\author[2\dagger]{Soo Ye Kim}
\affiliation[1]{Brown University}
\affiliation[2]{Adobe Research}
\contribution[*]{Work done during an internship at Adobe Research}
\contribution[\dagger]{Corresponding author}
\adobedata[Project Page]{\url{https://oliver-cong02.github.io/OmniXRef.github.io/}}

\begin{document}
\raggedbottom
\emergencystretch=3em
\begin{adobeabstract}
Creative intent in video editing can be expressed through text instructions (e.g., ``remove the people"), image references (e.g., subject appearance), video references (e.g., camera movement), or a combination of the above.
We introduce \frameworkname{}, a unified framework for multimodal multi-reference video editing that allows users to combine variable numbers of text, image, and video references to control diverse attributes of a source video.
We first devise a unified input design with a reference-bank interface and introduce \rropername{} to support generalization to more references.
Then our model, \modelname{}, uses a single-stream diffusion transformer to jointly process the source video and multimodal context from an MLLM in an interleaved sequence.
To address the scarcity of training data for video-reference editing, we construct \dataname{} using synthesis pipelines for different editing tasks and combinations of references.
We use curriculum learning to progressively increase task complexity and resolution.
To support systematic evaluation of video as a control modality, we introduce \benchname{}, a benchmark for video-reference attribute transfer across Camera, Motion, Content, Style, and Visual Effects (VFX).
Comprehensive evaluations show that our OmniXRef-Model outperforms task-specific baselines and commercial models.

\end{adobeabstract}

\begin{figure}[H]
\centering
\includegraphics[width=\textwidth]{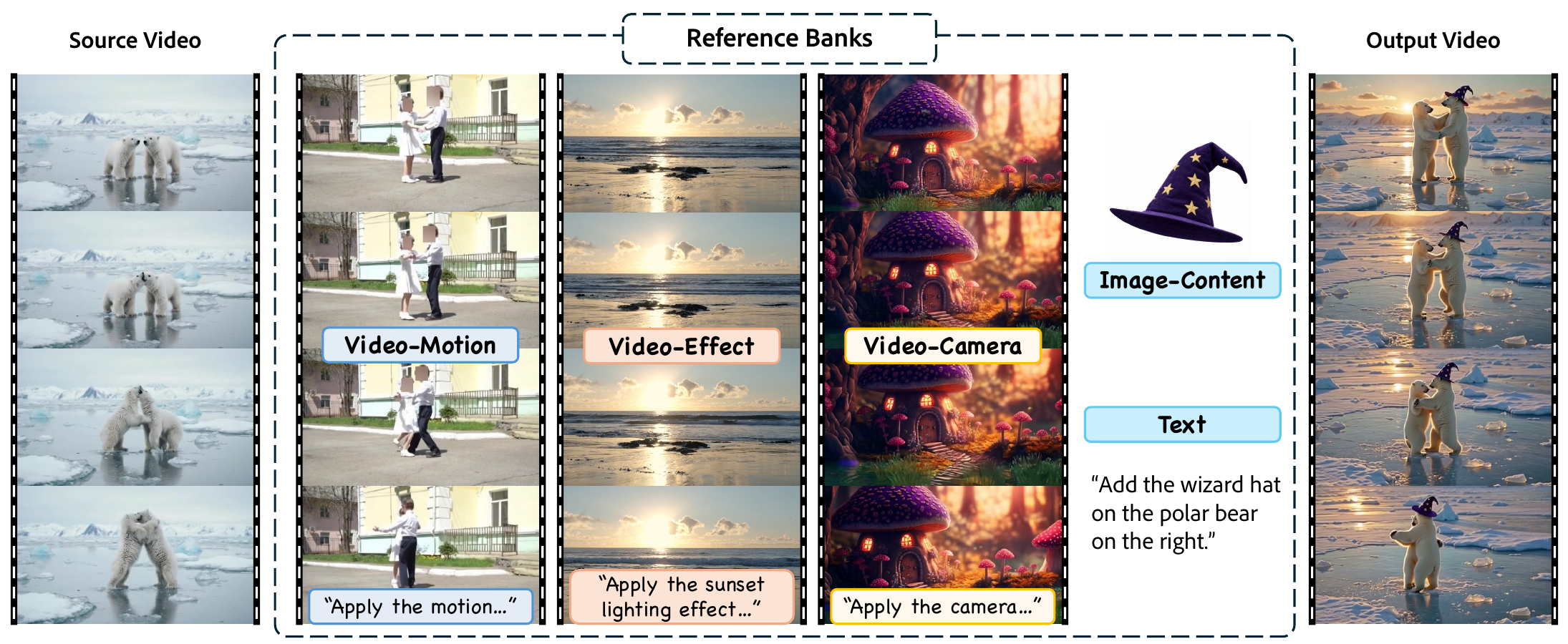}

\caption{\frameworkname{} supports multimodal (text, image and video) multi-reference video editing, transferring the desired attribute from each reference to the source video.}
\label{fig:teaser}
\end{figure}

\section{Introduction}
\label{sec:introduction}

Video editing plays a central role in modern digital media creation, enabling applications from film production and advertising to personalized visual storytelling.
While traditional video editing often requires time-consuming manual edits at each step of the editing pipeline, recent advances in generative models have made this more automatic. Given a source video, a video editing model aims to apply a specific edit guided by user intent (e.g., via a text prompt) while preserving everything else in the source video. However, describing intricate visual details and how they should dynamically change over time (e.g., camera, lighting, visual effects) can be highly challenging to specify through text and/or image guidance alone, leading to a need for a video reference. Going further, realizing a creative vision may require several multimodal references at once, each contributing a different aspect (e.g., object motion, visual effects, camera) of the intended result, as shown in Figure~\ref{fig:teaser}.

Despite this need, reference-guided video editing has mostly been studied in settings constrained by a single reference modality.
Existing methods support text instructions with an optional image reference~\citep{cong2025viva,mou2025instructx,ju2025editverse,lin2026kiwiedit}, multiple image references~\citep{liu2026rebind}, or a single video reference for a specific task such as camera control~\citep{luo2025camclonemaster} and visual effects (VFX)~\citep{jones2026refvfx}.
Frontier systems increasingly support broader combinations of multimodal references~\citep{bytedance2026seedance25,deepmind2026geminiomni,minimax2026h3}, though many only support references for generation~\citep{klingai2026video3,alibabacloud2026happyhorser2v,google2025veo31ingredients}, not editing.
In fact, a systematic formulation of \textit{multimodal multi-reference video editing} remains largely underdeveloped.

To this end, we introduce \emph{\frameworkname{}}, a unified framework for multimodal multi-reference video editing.
Our omni-reference formulation supports four dimensions: (i) number of references (1, ..., $K$), (ii) reference modality (text, image, video), (iii) reference attribute (Camera, Motion, Content, Style, VFX) and (iv) composition (text-image, text-video, image-video, text-image-video). The reference inputs are organized through a common reference-bank interface, with each bank specifying an editing task and instruction, together with an image or video reference when available, allowing \emph{\modelname{}} to combine multiple requested edits into one coherent output.
\modelname{} integrates an MLLM~\citep{bai2025qwen3vl} with a single-stream diffusion transformer (DiT), where the MLLM first encodes an interleaved sequence of the system prompt, source video and reference banks to derive the semantic context, then the DiT jointly processes this semantic context, source and reference VAE tokens, and noisy target-video tokens to perform the requested edits while preserving the rest of the source content. To reduce the computational cost of conditioning on multiple video references, we introduce Mix-Patchify (MP), which adapts different patchification granularity for each reference according to its editing task.
Furthermore, we introduce \rropername{} (RoPE with Reserved Reference Slots), which reserves a pool of positional slots for reference banks and randomly assigns banks to these slots during training, allowing support for more references at inference than seen in training.
We train \modelname{} with curriculum learning, progressively increasing task complexity and resolution to accelerate convergence.

Training \modelname{} across reference modalities requires large-scale data covering different reference counts and diverse editing tasks.
A key challenge is the limited availability of source--reference--target video triplets $(V_{\mathrm{src}}, V_{\mathrm{ref}}, V_{\mathrm{tgt}})$ covering a wide spectrum of editing tasks.
Thus, we develop data synthesis pipelines for single-video-reference editing across five categories: Camera, Motion, Content, Style, and VFX, as well as a compositional data synthesis pipeline to support multi-reference editing. These are assembled with text-only and text-image data, resulting in \emph{\dataname{}} containing $\sim$10M training examples.
Moreover, to address the absence of comprehensive video-reference editing benchmarks, we introduce \emph{\benchname{}}, comprising 125 cases across the five categories. The evaluation protocol measures how faithfully models follow the requested edit and preserve unedited source content, along with the visual quality and temporal consistency of the output.
Evaluations on \benchname{} show that \modelname{} outperforms task-specific baselines in their respective domains and demonstrates better reference adherence and source preservation compared to strong commercial models such as Seedance 2.5 and MiniMax H3.

In summary, our contributions are fourfold:
\begin{itemize}
    \item We introduce \emph{\frameworkname{}}, a unified framework that formulates multimodal multi-reference video editing along four dimensions: reference number, modality, attribute, and composition.
    Our proposed reference-bank interface enables flexible combinations of references, with great potential for training at scale.
    \item We develop \emph{\modelname{}}, combining \rropername{} to support generalization to larger reference sets and Mix-Patchify to reduce the computational cost of conditioning on multiple video references.
    \item We construct \emph{\dataname{}} with data synthesis pipelines for each single-video-reference editing task as well as compositions of two visual references.
    \item We introduce \emph{\benchname{}} to systematically evaluate reference adherence, visual quality, and source preservation, advancing future work towards video as a control modality for editing. Comprehensive evaluations show that our \modelname{} outperforms task-specific baselines and commercial models.
\end{itemize}

\section{Related Work}
\label{sec:related-work}

\paragraph{Video Generation and Editing.}
Video generation is evolving toward comprehensive content-creation systems for multimodal generation and editing~\citep{bytedance2026seedance25, deepmind2026geminiomni, klingteam2025klingomni, minimax2026h3}.
A central research direction is to expand the types of input that users can provide to specify an edit.
In this paper, we denote each editing setting by its reference inputs: $\mathrm{T}'$, $\mathrm{I}'$, and $\mathrm{V}'$ indicate a text, image, and video reference, respectively, and $\mathrm{V2V}$ denotes source-to-output video editing.
EditVerse~\citep{ju2025editverse}, VIVA~\citep{cong2025viva}, InstructX~\citep{mou2025instructx}, and Kiwi-Edit~\citep{lin2026kiwiedit} support \editsetting{T} and \editsetting{I}.
ReBind~\citep{liu2026rebind} and TIDE~\citep{liu2026tide} support multiple reference images and associate each with its intended edit.
Video references provide temporal information, including camera trajectories, dynamic lighting, and evolving visual effects (VFX), that image(s) cannot demonstrate and text alone may not specify precisely.
CamCloneMaster~\citep{luo2025camclonemaster} and RefVFX~\citep{jones2026refvfx} study camera control and VFX transfer in the \editsetting{V} setting, respectively.
Recent work~\citep{zhang2026omnitransfer,tan2025omnivideo,yang2026omnivideo2,wang2026vega,qin2026univigu,wei2025univideo,chen2026vino,pan2026omniweaving,wu2026loomvideo} integrates multimodal understanding into video generation and editing through MLLM-based conditioning or joint modeling to handle complex compositions of instructions and references.
Building on this progression, \frameworkname{} provides a unified formulation in which each text, image, or video reference is associated with an editing task and instruction, and multiple references can jointly specify the output.

\paragraph{Video Editing Datasets and Benchmarks.}
Training a general video editing model requires millions of high-quality data.
Ditto-1M~\citep{bai2025ditto} provides an open-sourced million-scale \editsetting{T} dataset.
RefVIE~\citep{lin2026kiwiedit} and RefVideo-6M~\citep{zi2026refvideo6m} are highlighted for large-scale \editsetting{I} data.
Specialized \editsetting{V} datasets accompany camera and VFX transfer methods~\citep{luo2025camclonemaster,jones2026refvfx}.
However, large-scale source--reference--target triplets spanning diverse \editsetting{V} tasks and their compositions remain scarce.
To address this gap, we construct \dataname{} with data synthesis pipelines for each single-video-reference editing task as well as compositions of two visual references.
For evaluation, ReCo-Bench~\citep{zhang2025reco} and VEFX-Bench~\citep{gao2026vefx} evaluate \editsetting{T}, EditVerseBench~\citep{ju2025editverse} and RefVIE-Bench~\citep{lin2026kiwiedit} cover \editsetting{T} and \editsetting{I}, and IntelligentVBench~\citep{pan2026omniweaving}, OmniVideo-1.0~\citep{klingteam2025klingomni}, and TIDE-Bench~\citep{liu2026tide} extend evaluation to multimodal composition.
However, none of them systematically evaluates transferring dynamic attributes from reference videos to a source video while preserving the remaining source content.
\benchname{} fills this gap by systematically evaluating reference adherence, visual quality, and source preservation, advancing future work towards video as a control modality for editing.

\section{\modelname{}}
\label{sec:method}

\subsection{Framework Overview}
\label{sec:unified-formulation}

\begin{figure}[t]
\centering
\includegraphics[width=0.95\linewidth]{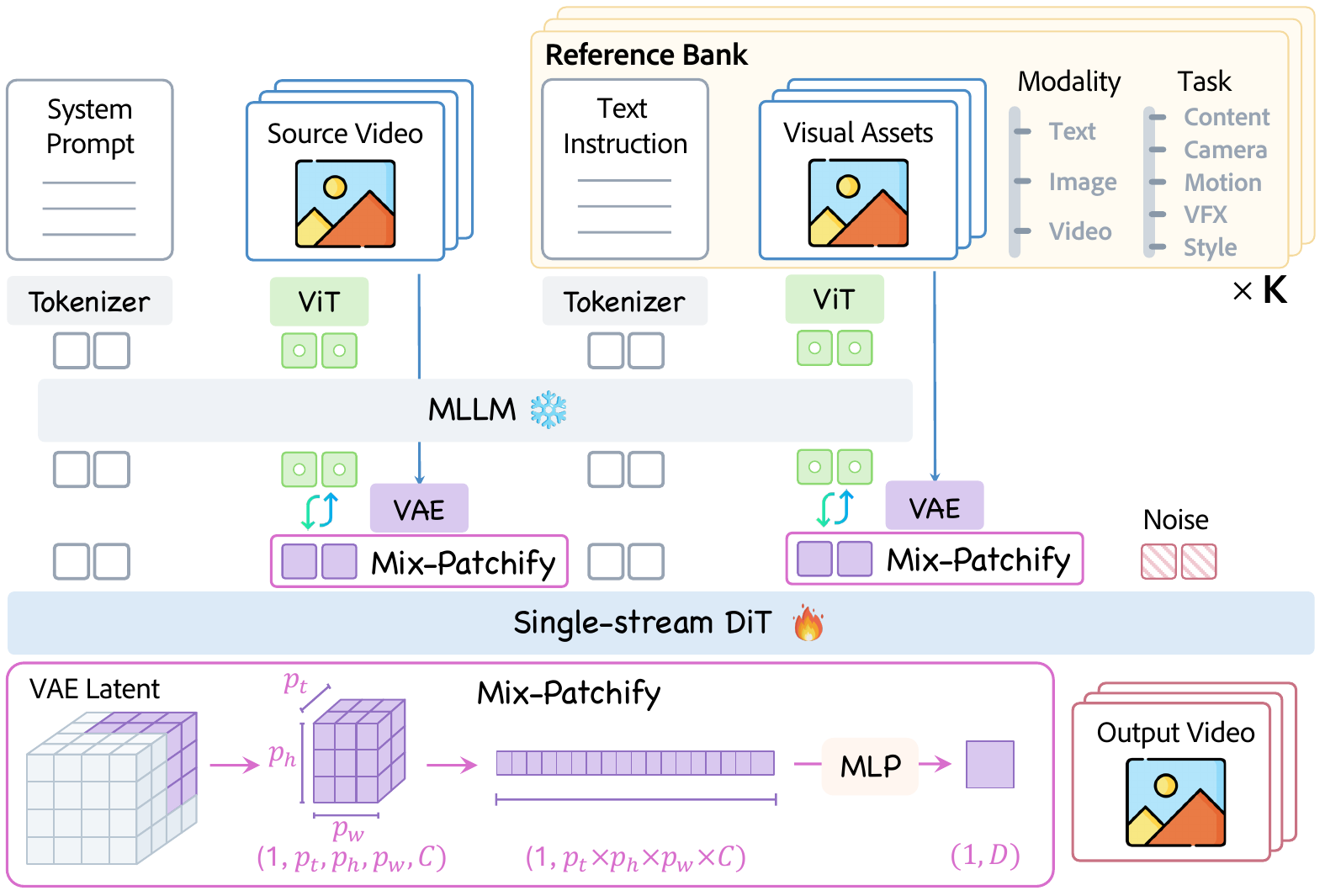}

\caption{\textbf{Overview of \modelname{}.}
Reference banks provide a unified interface for text, image, and video references, enabling multimodal multi-reference video editing.
An MLLM extracts semantic context from the interleaved inputs.
A single-stream DiT jointly processes the contextual text tokens, visual tokens, and noise through full self-attention, allowing multiple reference banks to condition the edited output simultaneously.}
\label{fig:pipeline}
\end{figure}

Coordinating multiple multimodal references within a single output requires a comprehensive understanding and reasoning of the whole input sequence.
Hence, as shown in Figure~\ref{fig:pipeline}, our \modelname{}, $F_{\theta}$, first uses an MLLM ($G$), which is a frozen Qwen3-VL-8B-Instruct~\citep{bai2025qwen3vl} in practice, to encode an interleaved input sequence and derive its semantic context, then employs a pretrained single-stream video diffusion transformer (DiT), whose architecture is similar to EditVerse~\citep{ju2025editverse}, to generate the edited video.

For a unified formulation, we represent text, image, and video references through a unified reference-bank interface.
Each bank consists of a task definition, an edit instruction, and an optional image or video reference:
\begin{equation}
B_i = \big[\underbrace{T_i^{\mathrm{task}}}_{\text{Task definition}};\ \underbrace{T_i^{\mathrm{edit}}}_{\text{Edit instruction}};\ \underbrace{V_i}_{\text{Optional visual reference}}\big].
\label{eq:reference-bank}
\end{equation}
For $i=1,\ldots,K$, $T_i^{\mathrm{task}}$ defines the scope of the editing task, $T_i^{\mathrm{edit}}$ specifies the concrete edit instruction, and $V_i$ is the reference image or video.
For a text-only bank, we set $T_i^{\mathrm{task}}=\emptyset$, $V_i=\emptyset$.

In \modelname{}, a semantic context $\mathcal{X}$ is first obtained by:
\begin{equation}
\mathcal{X}=G(T_{\mathrm{sys}}, T_{\mathrm{src}}, \mathcal{E}_{\mathrm{ViT}}(V_{\mathrm{src}}), \{B_i\}_{i=1}^{K}, T_{\mathrm{out}}),
\label{eq:mllm}
\end{equation}
where $T_{\mathrm{sys}}$ is the system prompt that defines the overall reference-guided editing task, $T_{\mathrm{src}}$ is the source marker \textit{``This is the source video to edit:''}, $T_{\mathrm{out}}$ is the output marker \textit{``The output video is:''}, $V_{\mathrm{src}}$ is the source video, and $\mathcal{E}_{\mathrm{ViT}}$ is a ViT encoder. Note that $V_i$ in $B_i$ also goes through $\mathcal{E}_{\mathrm{ViT}}$.

Then, the DiT, $H_{\theta}$, generates the output $H_{\theta}(\mathcal{X}', V_{\mathrm{tgt}}; \theta)$, where $\mathcal{X}'$ is the semantic context after replacing all visual tokens in $\mathcal{X}$ to $\mathcal{E}_{\mathrm{VAE}}(V)$, $\mathcal{E}_{\mathrm{VAE}}$ is the VAE encoder, and $V_{\mathrm{tgt}}$ is the ground truth edited video, which is noised during training. The DiT processes this sequence with full self-attention and denoises the noise tokens in context to generate the edited video.

\subsection{Mix-Patchify}
\label{sec:mix-patchify}
The patchification layer groups the VAE latents of a visual input into non-overlapping patches of size $p=(p_t,p_h,p_w)$ along the temporal, height, and width axes, and maps each flattened patch to a DiT token $\mathbf{e}_j$ with an MLP specific to $p$.
The pretrained DiT uses $p=(1,1,1)$ for fine-grained denoising.
We retain this granularity for the source and target tokens to preserve detailed source conditioning.
Reference tokens, however, do not all require such fine spatial granularity.
We therefore introduce \emph{Mix-Patchify} (MP), which represents each reference at a spatial granularity suited to its editing task while keeping $p_t=1$ throughout to preserve fine-grained temporal resolution for dynamic attributes. More details are provided in Appendix~\ref{app:mix-patchify}.
For a patch size $p$, MP reduces the token count by up to a factor of $p_t \cdot p_h \cdot p_w$.
This substantially lowers training and inference costs (Section~\ref{sec:inference-cost}), particularly with multiple references, while maintaining performance and facilitating convergence.

\subsection{\rropername}
\label{sec:r2-rope}
Multimodal multi-reference video editing data become increasingly difficult to construct as the number of references increases, fundamentally restricting the model's ability to compose more references at inference.
However, we attribute this limitation partly to a training--inference gap in the default sequential asset indexing of the 4D RoPE.
By default, the 4D RoPE coordinate of token $\mathbf{e}_j$ is $\mathbf{q}_{j}=(y_{j},x_{j},\tau_{j},\rho_s)$, where $(y_{j},x_{j},\tau_{j})$ denote its height, width, and temporal coordinates, and $\rho_s$ is the sequential index of the asset to which the token belongs with regards to the full input sequence.
Therefore, during training, the range of $\rho_s$ is bounded by the maximum number of assets in a training sequence.
In such a case, when the number of references at inference exceeds those seen during training, the model encounters out-of-distribution RoPE indices and relative offsets, hindering generalization.

To bypass this limitation, we propose \rropername{} (RoPE with Reserved Reference Slots), which modifies the fourth axis of the default 4D RoPE.
Specifically, we reserve fixed slots for the system prompt, the source video, and the noise tokens, together with a dedicated pool of slots for reference banks, where the size of this pool is larger than the number of references supported with training data.
In practice, we reserve eight slots for reference banks.
All tokens belonging to the reference bank $B_i$ share the same fourth-axis coordinate $\rho_i$. The key is to randomly sample from the reserved reference slots during training. When a training example contains multiple references, non-overlapping slots are selected such that different banks receive distinct coordinates.
As this design exposes every reserved reference slot to a reference input during training, additional references at inference time remain within the RoPE range seen during training, i.e., becomes \textit{in-distribution}.
\rropername{} thus avoids the training--inference gap and is able to support more number of references at inference time without requiring training examples for every reference number and combination.

\subsection{Curriculum Learning}
\label{sec:curriculum-learning}
We train \modelname{} in three stages that progressively increase task complexity and output resolution.

\noindent\textbf{Stage 1: Single-reference adaptation.}
We begin at 272P with a mixture of \editsetting{T}, \editsetting{I}, and \editsetting{V} examples.
This stage adapts the pretrained model to the unified reference-bank interface and establishes task-specific transfer ability while keeping computational cost low.
The learning rate starts at $1.5\times10^{-5}$ and decays to $1\times10^{-5}$ over the first $5{,}000$ steps.

\noindent\textbf{Stage 2: Multi-reference composition.}
At 272P, we retain the Stage~1 tasks and introduce \editsetting{IV} and \editsetting{VV} examples that combine two visual references.
The model learns to use multiple \rropername{} slots together, providing supervision for compositions beyond any single reference bank.
The learning rate remains at $1\times10^{-5}$.

\noindent\textbf{Stage 3: High-resolution fine-tuning.}
We increase the resolution of all training examples to 480P while keeping the Stage~2 mixture and other settings unchanged.
The stage-wise sampling weights are reported in Table~\ref{tab:data-mixture}.

\section{Data}
\label{sec:data}
We assemble \dataname{} to train \modelname{} across reference modalities, numbers of references, editing tasks, and attribute combinations, resulting in $\sim$10M training examples spanning text-only (\editsetting{T}), one image or video reference ($(\mathrm{I}/\mathrm{V})'\mathrm{V2V}$), and two visual references (\editsetting{IV} or \editsetting{VV}).
\begin{wraptable}{R}{0.48\textwidth}
\centering
\footnotesize
\setlength{\tabcolsep}{2.5pt}
\renewcommand{\arraystretch}{1.05}
\begin{tabular*}{\linewidth}{@{}l@{\extracolsep{\fill}}ccr@{}}
\toprule
\textbf{Task} & \multicolumn{2}{c}{\textbf{Weight}} & \textbf{\#Samples} \\
\cmidrule(lr){2-3}
 & Stage 1 & Stage 2/3 & \\
\midrule
\multicolumn{4}{@{}l}{\textit{\editsetting{T} --- Text-only}} \\
\midrule
General & 20\% & 20\% & 7M \\
\midrule
\multicolumn{4}{@{}l}{\textit{$(\mathrm{I}/\mathrm{V})'\mathrm{V2V}$ --- Single visual reference}} \\
\midrule
Camera & 16\% & 8\% & 1.5M \\
Motion & 24\% & 12\% & 100K \\
Content & 16\% & 8\% & 650K \\
VFX & 16\% & 8\% & 100K \\
Style & 8\% & 4\% & 650K \\
\midrule
\multicolumn{4}{@{}l}{\textit{\editsetting{IV} / \editsetting{VV} --- Two visual references}} \\
\midrule
Content + Camera & 0\% & 20\% & 50K \\
Motion + Camera & 0\% & 20\% & 50K \\
\midrule
\textbf{Total} & \textbf{100\%} & \textbf{100\%} & \textbf{10M} \\
\bottomrule
\end{tabular*}
\caption{Statistics of \dataname{} with the sampling weight of each task at each curriculum stage.}
\label{tab:data-mixture}
\end{wraptable}

\noindent\textbf{Text-only editing.}
Our text-only data contains $\sim$7M high-quality in-house source--target pairs with diverse instructions.
We retain this supervision throughout training to maintain general editing foundations, including instruction following and source video preservation.
It also helps prevent the model from overfitting to a particular reference modality.

\noindent\textbf{Visual-reference data.}
We construct source--reference--target triplets $(V_{\mathrm{src}},V_{\mathrm{ref}},V_{\mathrm{tgt}})$ such that the source and target differ primarily in the edited attribute, while the reference and target share that attribute but depict a distinct content.
This design discourages the model from directly copying reference content instead of transferring the intended attribute.
We use Qwen3-VL-8B-Instruct~\citep{bai2025qwen3vl} as the MLLM throughout data construction.
(1) For \textbf{VFX} and \textbf{Style}, which specify an evolving visual effect and a global appearance change such as stylization or dynamic lighting, respectively, we use a shared synthesis pipeline.
The MLLM filters and captions a large-scale text-to-video collection, identifying $\sim$2M candidate target videos with effects such as flames, explosions, smoke, and particles, or with distinctive styles.
For each target $V_{\mathrm{tgt}}$, GPT Image 2~\citep{openai2026gptimage2} is used to remove the detected attribute (e.g., smoke) from its first frame using the instruction formulated by the MLLM.
Next, a video propagation model~\citep{liu2025genprop} propagates this first-frame edit through $V_{\mathrm{tgt}}$ to produce $V_{\mathrm{src}}$, removing the attribute throughout the video while preserving its content.
Then for $V_{\mathrm{ref}}$, MLLM describes a plausible different scene exhibiting the same attribute; GPT Image 2 generates the first frame using this prompt, and Wan2.2-TI2V-5B~\citep{wan2025wan22} generates a video from this first frame.

(2) For \textbf{Camera}, where the reference specifies a camera trajectory, framing, and timing, we sample two visually distinct videos, $V_1$ and $V_2$, from an in-house text-to-video collection.
We reshoot both with Vista4D~\citep{lin2026vista4d} under the same sampled camera path, obtaining $V_1^{\mathrm{cam}}$ and $V_2^{\mathrm{cam}}$ and thus two triplets: $(V_1,V_2^{\mathrm{cam}},V_1^{\mathrm{cam}})$ and $(V_2,V_1^{\mathrm{cam}},V_2^{\mathrm{cam}})$.
The paths include primitives like orbits, dollies, cranes, optical zooms, pans and tilts, and freeform movements with varied scale, speed, framing, and rhythm.

(3) For \textbf{Motion}, where the reference specifies an object (e.g., human) pose sequence, we start from paired videos $(V_1,V_2)$ in MotionPair-60K~\citep{yan2026scail2}, in which different characters perform the same motion.
For each character, Wan2.2-TI2V-5B generates a video with different motion, yielding $V_1^{\mathrm{alt}}$ and $V_2^{\mathrm{alt}}$.
We form $(V_1^{\mathrm{alt}},V_2,V_1)$ and $(V_2^{\mathrm{alt}},V_1,V_2)$, so that the reference demonstrates the target motion through a different character while the source retains the target character with a different motion.
(4) For \textbf{Content}, we mix RefVIE~\citep{lin2026kiwiedit}, which already provides image-reference triplets $(V_{\mathrm{src}},I_{\mathrm{ref}},V_{\mathrm{tgt}})$ for inserting, removing, or replacing subjects and backgrounds, as well as video harmonization triplets in which a video reference supplies a dynamic background for a source foreground.

\noindent\textbf{Compositional references.}
As the model cannot learn to \textit{combine} multiple-reference attributes into one output only by learning from single-reference data,
we obtain compositional data by applying the Camera reshooting pipeline to Content and Motion data pipelines, producing \editsetting{IV} and \editsetting{VV} data.

After MLLM quality filtering, the statistics of retained data are summarized in Table~\ref{tab:data-mixture}.

\section{Experiments}
\label{sec:experiments}
\subsection{Evaluation Setup}
\label{sec:evaluation-setup}
\begin{table}[htbp]
\centering
\caption{Per-category VLM evaluation of video-reference attribute transfer on \benchname{}.}
\label{tab:video-reference-editing-full}
\begingroup
\newcommand{\commercialtext}[1]{\textcolor{black!45}{#1}}
\footnotesize
\setlength{\tabcolsep}{2.4pt}
\renewcommand{\arraystretch}{1.0}
\begin{tabular*}{\linewidth}{@{\extracolsep{\fill}}l*{12}{c}@{}}
\toprule
Model & \multicolumn{4}{c}{Content} & \multicolumn{4}{c}{Style} & \multicolumn{4}{c}{VFX}\\
\cmidrule(lr){2-5}\cmidrule(lr){6-9}\cmidrule(l){10-13}
 & RA & VQ & SP & Score & RA & VQ & SP & Score & RA & VQ & SP & Score\\
\midrule
\multicolumn{13}{@{}c@{}}{\textcolor{black!65}{\textsc{Open-source}}}\\
RefVFX & 0.94 & 3.35 & 4.55 & 1.98 & 1.12 & 3.55 & \textbf{4.90} & 2.24 & 1.69 & 3.40 & 4.39 & 2.49\\
\midrule
\multicolumn{13}{@{}c@{}}{\textcolor{black!65}{\textsc{Ours}}}\\
\textbf{\frameworkname{} (MiniMax-H3)} & 4.22 & 3.61 & 4.08 & 3.70 & \textbf{3.75} & \textbf{3.74} & 4.40 & 3.55 & \textbf{3.56} & \textbf{3.63} & \textbf{4.85} & \textbf{3.83}\\
\textbf{\frameworkname{}} & \textbf{4.69} & \textbf{3.94} & \textbf{4.67} & \textbf{4.49} & 3.71 & 3.60 & 4.65 & \textbf{3.75} & \textbf{3.56} & 3.25 & 4.52 & 3.55\\
\midrule\midrule
\multicolumn{13}{@{}c@{}}{\textcolor{black!65}{\textsc{Commercial}}}\\
\commercialtext{MiniMax-H3} & \commercialtext{3.81} & \commercialtext{4.03} & \commercialtext{1.97} & \commercialtext{2.23} & \commercialtext{2.92} & \commercialtext{3.92} & \commercialtext{1.75} & \commercialtext{0.94} & \commercialtext{3.09} & \commercialtext{3.56} & \commercialtext{1.90} & \commercialtext{1.32}\\
\commercialtext{Seedance 2.0} & \commercialtext{4.69} & \commercialtext{3.93} & \commercialtext{4.03} & \commercialtext{3.73} & \commercialtext{3.71} & \commercialtext{3.85} & \commercialtext{3.95} & \commercialtext{3.22} & \commercialtext{3.69} & \commercialtext{3.72} & \commercialtext{3.74} & \commercialtext{2.92}\\
\commercialtext{Seedance 2.5} & \commercialtext{4.81} & \commercialtext{3.76} & \commercialtext{4.25} & \commercialtext{4.04} & \commercialtext{3.52} & \commercialtext{3.47} & \commercialtext{3.02} & \commercialtext{2.30} & \commercialtext{3.78} & \commercialtext{3.50} & \commercialtext{2.84} & \commercialtext{2.22}\\
\bottomrule
\addlinespace[1.0em]
\toprule
Model & \multicolumn{4}{c}{Camera} & \multicolumn{4}{c}{Motion} & \multicolumn{4}{c}{Overall avg.}\\
\cmidrule(lr){2-5}\cmidrule(lr){6-9}\cmidrule(l){10-13}
 & RA & VQ & SP & Score & RA & VQ & SP & Score & RA & VQ & SP & Score\\
\midrule
\multicolumn{13}{@{}c@{}}{\textcolor{black!65}{\textsc{Open-source}}}\\
CamCloneMaster & 2.52 & 3.74 & 4.99 & 3.20 & -- & -- & -- & -- & -- & -- & -- & --\\
MotionFollower & -- & -- & -- & -- & 2.02 & 2.43 & 2.99 & 2.14 & -- & -- & -- & --\\
\midrule
\multicolumn{13}{@{}c@{}}{\textcolor{black!65}{\textsc{Ours}}}\\
\textbf{\frameworkname{} (MiniMax-H3)} & \textbf{3.75} & \textbf{3.92} & 4.91 & \textbf{3.99} & 2.14 & \textbf{3.92} & \textbf{4.80} & 2.88 & 3.48 & \textbf{3.76} & 4.61 & 3.59\\
\textbf{\frameworkname{}} & 3.48 & 3.74 & \textbf{5.00} & 3.81 & \textbf{2.92} & 3.87 & 4.56 & \textbf{3.30} & \textbf{3.67} & 3.68 & \textbf{4.68} & \textbf{3.78}\\
\midrule\midrule
\multicolumn{13}{@{}c@{}}{\textcolor{black!65}{\textsc{Commercial}}}\\
\commercialtext{MiniMax-H3} & \commercialtext{2.44} & \commercialtext{3.89} & \commercialtext{3.67} & \commercialtext{2.07} & \commercialtext{0.05} & \commercialtext{3.82} & \commercialtext{2.23} & \commercialtext{0.69} & \commercialtext{2.46} & \commercialtext{3.84} & \commercialtext{2.30} & \commercialtext{1.45}\\
\commercialtext{Seedance 2.0} & \commercialtext{2.65} & \commercialtext{3.81} & \commercialtext{4.97} & \commercialtext{3.27} & \commercialtext{1.85} & \commercialtext{3.84} & \commercialtext{4.18} & \commercialtext{2.45} & \commercialtext{3.32} & \commercialtext{3.83} & \commercialtext{4.17} & \commercialtext{3.12}\\
\commercialtext{Seedance 2.5} & \commercialtext{2.33} & \commercialtext{3.89} & \commercialtext{4.98} & \commercialtext{3.07} & \commercialtext{1.58} & \commercialtext{3.86} & \commercialtext{3.45} & \commercialtext{2.00} & \commercialtext{3.21} & \commercialtext{3.70} & \commercialtext{3.71} & \commercialtext{2.73}\\
\bottomrule
\end{tabular*}
\endgroup
\end{table}

\noindent\textbf{Benchmarks.}
To address the absence of comprehensive video-reference editing benchmarks, we introduce \benchname{}, which contains 125 samples across five categories: Camera, Motion, Content, Style, and VFX.
Each sample pairs a source video with a reference video and specifies which attribute to transfer.
We~use~Gemini~3.8~Flash~\citep{google2026gemini38flash} as a VLM judge with a systematic evaluation prompt designed for each category;
Details are deferred to Appendix~\ref{app:benchmark-evaluation}.
Additionally, we also report ReCo-Bench~\citep{zhang2025reco} and VEFX-Bench~\citep{gao2026vefx} to evaluate whether \modelname{} preserves the foundational \editsetting{T} capability.

\noindent\textbf{Baselines.}
Few existing methods support multimodal multi-reference video editing, so we mainly compare with the commercial models Seedance~2.0~\citep{bytedance2026seedance20}, Seedance~2.5~\citep{bytedance2026seedance25}, and MiniMax-H3~\citep{minimax2026h3}.
We also compare with task-specific expert models: CamCloneMaster~\citep{luo2025camclonemaster} for Camera, MotionFollower~\citep{tu2025motionfollower} for Motion, and RefVFX~\citep{jones2026refvfx} for Content, Style, and VFX.
On ReCo-Bench and VEFX-Bench, we compare with the methods on their leaderboards.
In all tables, commercial systems are shown in grey for reference, and boldface marks the best result among the remaining methods.

\noindent\textbf{Metrics.}
For \benchname{}, the VLM judge scores three aspects: Reference Adherence (RA), whether the output correctly transfers the requested attribute from the reference; Visual Quality (VQ), the visual and temporal quality of the output; and Source Preservation (SP), whether the remaining source content is preserved without importing unrelated reference content.
Following CamI2V~\citep{zheng2024cami2v}, we additionally measure trajectory accuracy for the Camera reference, reporting Rotation Distance (RotErr), Translation Error (TransErr), and Camera Motion Consistency (CamMC).
For ReCo-Bench and VEFX-Bench, we follow their official evaluation suites.

\subsection{Quantitative Comparison}
\label{sec:main-results}
Table~\ref{tab:video-reference-editing-full} reports RA, VQ, SP, and the composite Score for each category of \benchname{}, together with the five-category average of each metric; dashes mark unavailable results.
\modelname{} achieves the highest overall Score, RA, and SP, leads on Motion, Content, and Style, and outperforms task-specific baselines in their respective categories and commercial systems overall.
The remaining VQ gap relative to commercial systems may reflect differences in pretrained backbone capacity; nevertheless, our model designs deliver strong reference adherence and source preservation while maintaining competitive visual quality.
To~assess~whether~our~unified~formulation generalizes beyond the pretrained DiT used in \modelname{}, we fine-tune MiniMax-H3 on \dataname{} starting from its Ref2VA pretrained checkpoint, without modifying its architecture or adding task-specific components.
We refer to the resulting model as \frameworkname{} (MiniMax-H3).
It improves on the original MiniMax-H3 across all five categories and surpasses \modelname{} on Camera and VFX, supporting the portability of our formulation and the practical value of \dataname{}.
These results also suggest that more tailored adaptation strategies could yield further gains.
Geometry-based camera trajectory errors further support the camera adherence of \modelname{} (Table~\ref{tab:camera-trajectory}), while results on VEFX-Bench and ReCo-Bench show that it retains competitive \editsetting{T} capability (Appendix~\ref{app:additional-results}).
\subsection{Qualitative Comparison}
\label{sec:qualitative}
\begin{figure}[ht]
\centering
\includegraphics[width=0.99\linewidth]{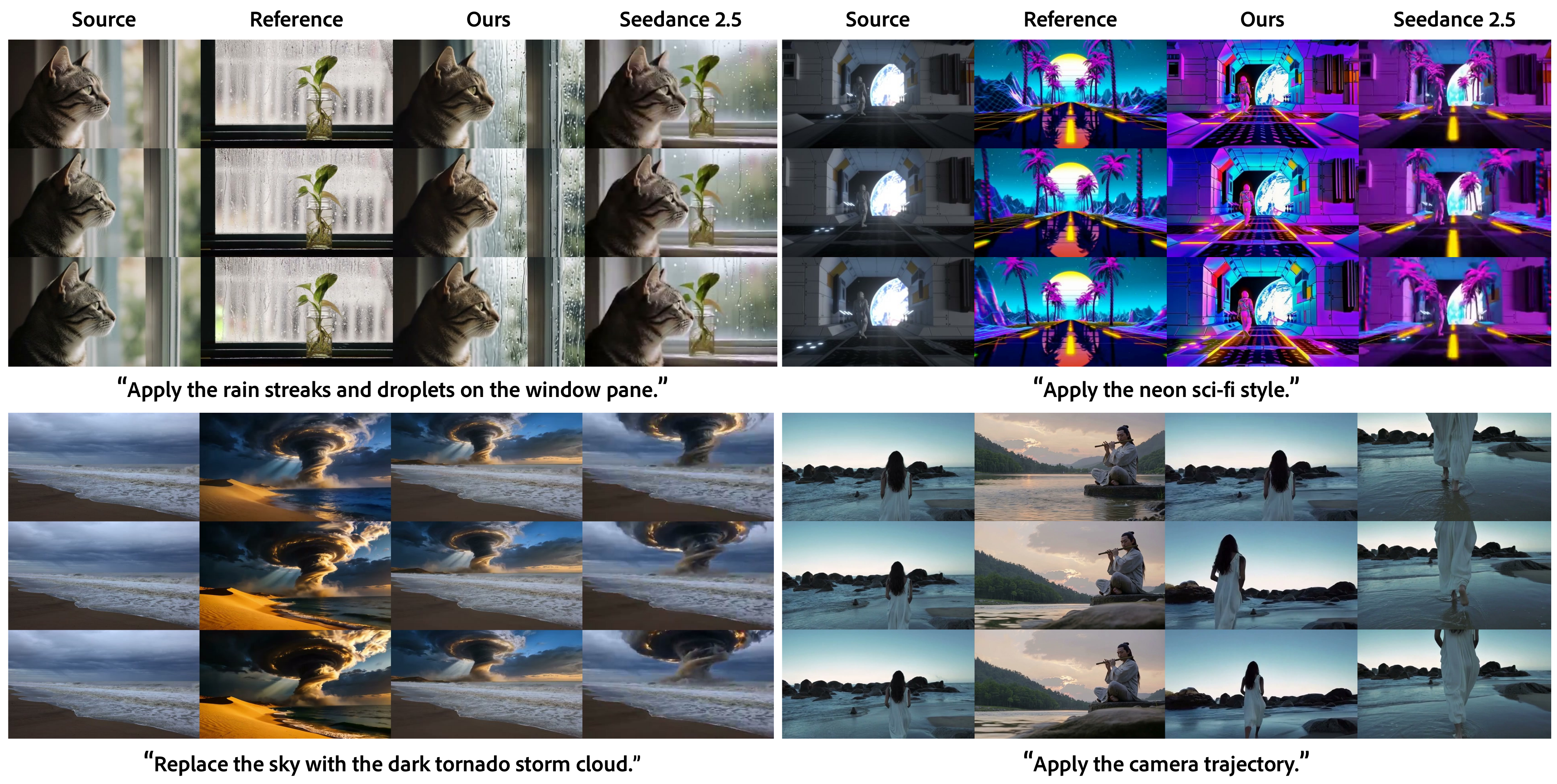}

\caption{Qualitative comparison of \modelname{} and Seedance~2.5 on \editsetting{V} across visual effects, style, content, and camera editing.}
\label{fig:qualitative-comparison}
\end{figure}

Figure~\ref{fig:qualitative-comparison} compares \modelname{} with Seedance~2.5 on four \editsetting{V} cases spanning VFX, Style, Content, and Camera.
Consistent with Table~\ref{tab:video-reference-editing-full}, \modelname{} shows clear advantages in reference adherence and source preservation.
For example, in the top-left sample in Figure~\ref{fig:qualitative-comparison}, Seedance~2.5 responds to the rain but does not reproduce the droplets shown in the reference.
It also alters the window frame, failing to preserve unedited source content, and even imports the potted plant from the reference, a form of reference leakage that also occurs in the top-right sample.
In contrast, \modelname{} precisely transfers the reference rain streaks and droplets onto the window pane while keeping the rest of the scene unchanged and avoiding reference leakage.

\FloatBarrier
\subsection{Camera Trajectory Evaluation}
\label{sec:camera-trajectory-results}
\begin{wraptable}{R}{0.48\textwidth}
\centering
\caption{Camera trajectory errors on \benchname{}, summed over frames; RotErr is in radians.}
\label{tab:camera-trajectory}
\begingroup
\newcommand{\commercialtext}[1]{\textcolor{black!45}{#1}}
\footnotesize
\setlength{\tabcolsep}{2pt}
\renewcommand{\arraystretch}{1.15}
\begin{tabular*}{\linewidth}{@{}l@{\extracolsep{\fill}}ccc@{}}
\toprule
Model & RotErr$\downarrow$ & TransErr$\downarrow$ & CamMC$\downarrow$\\
\midrule
CamCloneMaster & 2.086 & 9.329 & 10.332\\
\midrule
\textbf{\frameworkname{}} & \textbf{0.874} & \textbf{3.961} & \textbf{4.437}\\
\midrule\midrule
\multicolumn{4}{@{}c@{}}{\textcolor{black!65}{\textsc{Commercial}}}\\
\commercialtext{Seedance 2.0} & \commercialtext{4.443} & \commercialtext{22.455} & \commercialtext{24.008}\\
\commercialtext{Seedance 2.5} & \commercialtext{4.373} & \commercialtext{22.445} & \commercialtext{24.001}\\
\bottomrule
\end{tabular*}
\endgroup
\vspace{-6pt}
\end{wraptable}

To complement the VLM-based camera scores, we estimate camera poses independently for the reference and output videos with VGGT-$\Omega$~\citep{wang2026vggtomega} on 32 aligned frames.
\begin{table}[t]
\centering
\caption{Per-category ablations of \modelname{} on \benchname{}.}
\label{tab:ablations-full}
\begingroup
\footnotesize
\setlength{\tabcolsep}{2.4pt}
\renewcommand{\arraystretch}{1.0}
\begin{tabular*}{\linewidth}{@{\extracolsep{\fill}}l*{12}{c}@{}}
\toprule
Variant & \multicolumn{4}{c}{Content} & \multicolumn{4}{c}{Style} & \multicolumn{4}{c}{VFX}\\
\cmidrule(lr){2-5}\cmidrule(lr){6-9}\cmidrule(l){10-13}
 & RA & VQ & SP & Score & RA & VQ & SP & Score & RA & VQ & SP & Score\\
\midrule
w/o Mix-Patchify & 3.75 & 3.80 & 3.30 & 2.97 & 3.40 & 3.22 & 3.58 & 2.71 & \textbf{4.12} & \textbf{3.61} & 3.64 & 3.30\\
Reference-bank dropout & 3.06 & 3.73 & 3.30 & 2.48 & 3.25 & 3.50 & 4.45 & 3.29 & 3.72 & 3.47 & 4.14 & 3.42\\
w/o \rropername{} & 3.31 & 3.47 & 3.89 & 3.06 & 2.83 & \textbf{3.62} & 4.44 & 3.14 & 3.41 & 3.48 & \textbf{4.66} & \textbf{3.64}\\
w/o curriculum learning & 3.69 & 3.56 & 3.70 & 3.07 & 3.08 & 3.44 & 4.32 & 3.15 & 3.69 & 3.37 & 4.22 & 3.51\\
\midrule
\textbf{\frameworkname{} (full)} & \textbf{4.69} & \textbf{3.87} & \textbf{4.70} & \textbf{4.48} & \textbf{3.71} & 3.60 & \textbf{4.65} & \textbf{3.74} & 3.56 & 3.25 & 4.52 & 3.55\\
\bottomrule
\addlinespace[1.0em]
\toprule
Variant & \multicolumn{4}{c}{Camera} & \multicolumn{4}{c}{Motion} & \multicolumn{4}{c}{Overall avg.}\\
\cmidrule(lr){2-5}\cmidrule(lr){6-9}\cmidrule(l){10-13}
 & RA & VQ & SP & Score & RA & VQ & SP & Score & RA & VQ & SP & Score\\
\midrule
w/o Mix-Patchify & 3.01 & \textbf{3.87} & 4.67 & 3.24 & 1.51 & 3.86 & 3.92 & 2.11 & 3.16 & \textbf{3.67} & 3.82 & 2.86\\
Reference-bank dropout & 2.95 & 3.40 & 4.25 & 2.85 & 2.07 & 3.80 & 3.43 & 2.11 & 3.01 & 3.58 & 3.91 & 2.83\\
w/o \rropername{} & 3.20 & 3.69 & 4.94 & 3.61 & 1.57 & 3.74 & \textbf{4.60} & 2.44 & 2.86 & 3.60 & 4.50 & 3.18\\
w/o curriculum learning & 2.69 & 3.67 & 4.97 & 3.32 & 1.81 & \textbf{3.92} & 4.53 & 2.57 & 2.99 & 3.59 & 4.35 & 3.13\\
\midrule
\textbf{\frameworkname{} (full)} & \textbf{3.48} & 3.74 & \textbf{5.00} & \textbf{3.81} & \textbf{2.86} & 3.88 & 4.57 & \textbf{3.27} & \textbf{3.66} & \textbf{3.67} & \textbf{4.69} & \textbf{3.77}\\
\bottomrule
\end{tabular*}
\endgroup
\end{table}

We compare trajectories relative to their first frames and sum the errors over the remaining 31 frames; Appendix~\ref{app:camera-metrics} defines the rotation error (RotErr), translation error (TransErr), and camera motion consistency (CamMC).
Table~\ref{tab:camera-trajectory} shows that \modelname{} has the lowest error on all three measures among the evaluated open-source and commercial baselines.
\par\pagebreak[4]\noindent
Because translations are normalized independently for each trajectory, these errors capture motion direction, shape, and timing rather than absolute translation magnitude or changes in focal length.
\FloatBarrier

\subsection{Ablation Studies}
\label{sec:ablations}
We ablate the key designs of \modelname{}, including Mix-Patchify (Section~\ref{sec:mix-patchify}), \rropername{} (Section~\ref{sec:r2-rope}), curriculum learning (Section~\ref{sec:curriculum-learning}), and the dropout strategy, and report the per-category results in Table~\ref{tab:ablations-full}.

\noindent\textbf{Mix-Patchify.} Although applying $(1,1,1)$ patches to all references preserves the most complete reference information,
it provides no spatial prior: the model must learn on its own how to extract the requested attribute from long, information-dense reference sequences, which makes learning harder and slows convergence within the same number of training steps.
In contrast, Mix-Patchify uses task-specific patch sizes as a spatial prior, keeping fine-grained patches where appearance details matter (Content) and coarser patches where the attribute is mainly global or dynamic (Camera, Motion, Style, and VFX), which reduces the learning difficulty, making the model converge faster.

\noindent\textbf{\rropername{}.} Replacing \rropername{} with the default sequential asset indexing causes the smallest overall degradation among the ablations.
We attribute this degradation to an intrinsic indexing conflict: under sequential indexing, the fourth-axis index of the noise tokens shifts with the number of references, so for example, the same index denotes the output in samples with one reference but a reference in samples with more.
Mixing such samples during training therefore introduces conflicting positional signals that make learning harder.

\noindent\textbf{Curriculum learning.} Without curriculum learning, the model is trained at 480P on the full data pool.
Within the same number of training steps, it must acquire all capabilities at once without the inductive bias of easy-to-hard training, which hinders optimization and degrades performance.
Training at 480P throughout also makes each step considerably slower.

\noindent\textbf{Reference-bank dropout.} By default, we drop only the editing instruction $T_i^{\mathrm{edit}}$ for classifier-free guidance~\citep{ho2022classifierfree}, so that the unconditional prediction remains grounded in the same visual reference and the guidance amplifies only the attribute extraction specified by the text instruction.
In contrast, dropping the entire bank makes the guidance amplify the influence of the whole reference rather than only the requested attribute, which likely induces reference leakage and degrades performance.

\FloatBarrier

\subsection{Analysis of Reference Conditioning}
\label{sec:reference-analysis}
\begin{figure}[htbp]
\centering
\includegraphics[width=0.99\linewidth]{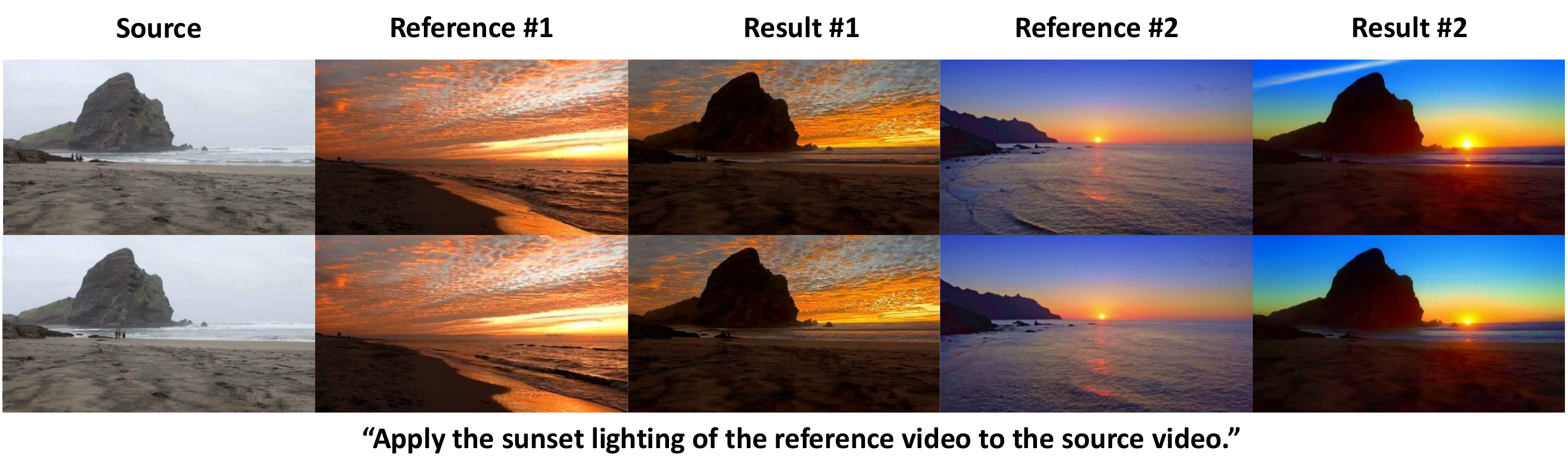}

\caption{Responsiveness to different visual references with a fixed source video and edit instruction.}
\label{fig:reference-responsiveness}
\end{figure}

\noindent\textbf{Responsiveness to visual references.}
\label{sec:reference-responsiveness}
To verify that \modelname{} actually uses the visual asset in the reference bank rather than relying on the editing instruction alone, we fix the source video and the instruction and only change the reference video.
In Figure~\ref{fig:reference-responsiveness}, two references with different sunset lighting are paired with the same instruction, ``Apply the sunset...''.
The two outputs reproduce the distinct lighting of their respective visual references, confirming that the visual asset determines the appearance while the instruction specifies which attribute to transfer.

\begin{figure}[!htb]
\centering
\includegraphics[width=0.99\linewidth]{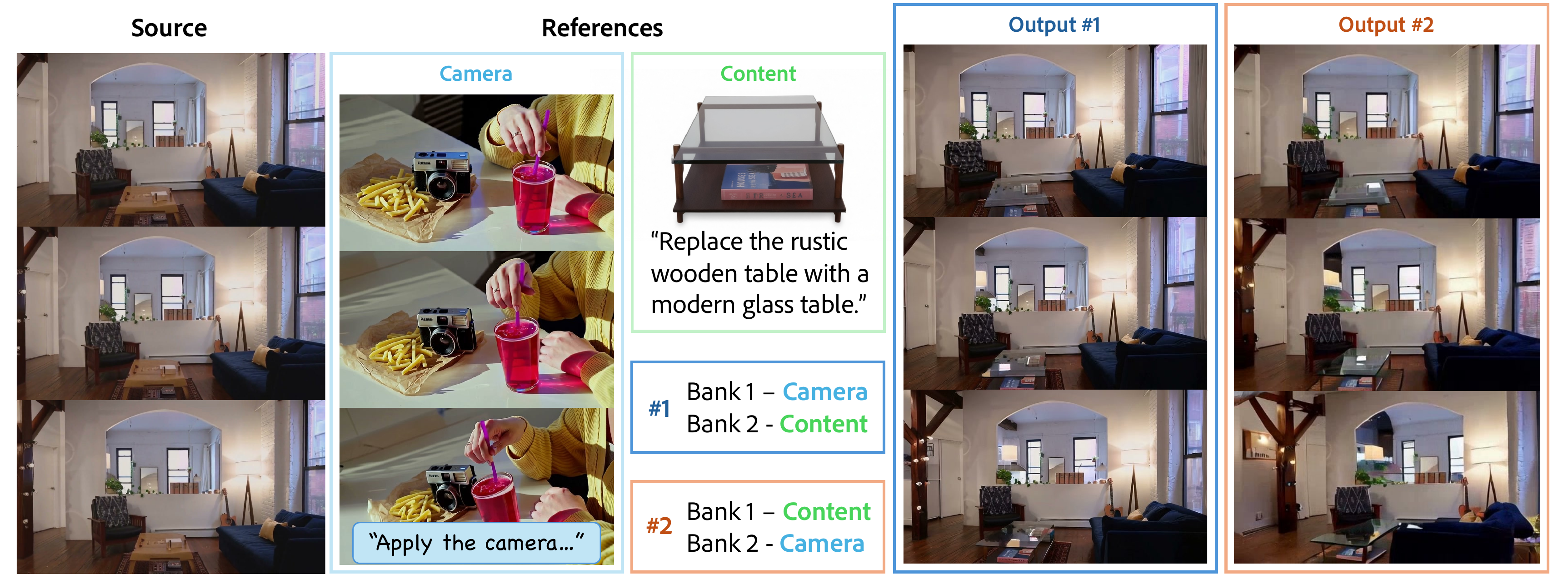}
\caption{Reference-order robustness when swapping reference bank slots in \rropername{}.}
\label{fig:reference-order-robustness}
\end{figure}

\noindent\textbf{Robustness to reference order.}
\label{sec:reference-order-robustness}
With multiple references, a natural question is whether the order of the reference banks, that is, which \rropername{} slot each bank occupies, affects the result.
To test this, we keep everything the same while swapping the two reference banks between their slots.
As shown in Figure~\ref{fig:reference-order-robustness}, both orders correctly transfer the camera motion and replace the table, indicating that \modelname{} does not collapse into associating particular slots with particular attributes.
This robustness stems from randomized slot assignment, which exposes every reserved slot to diverse references during training.
As a result, although each training sample contains at most two visual references, all reserved slots are activated without bias.

\begin{figure}[!htb]
\vspace{-6pt}
\centering
\includegraphics[width=0.99\linewidth]{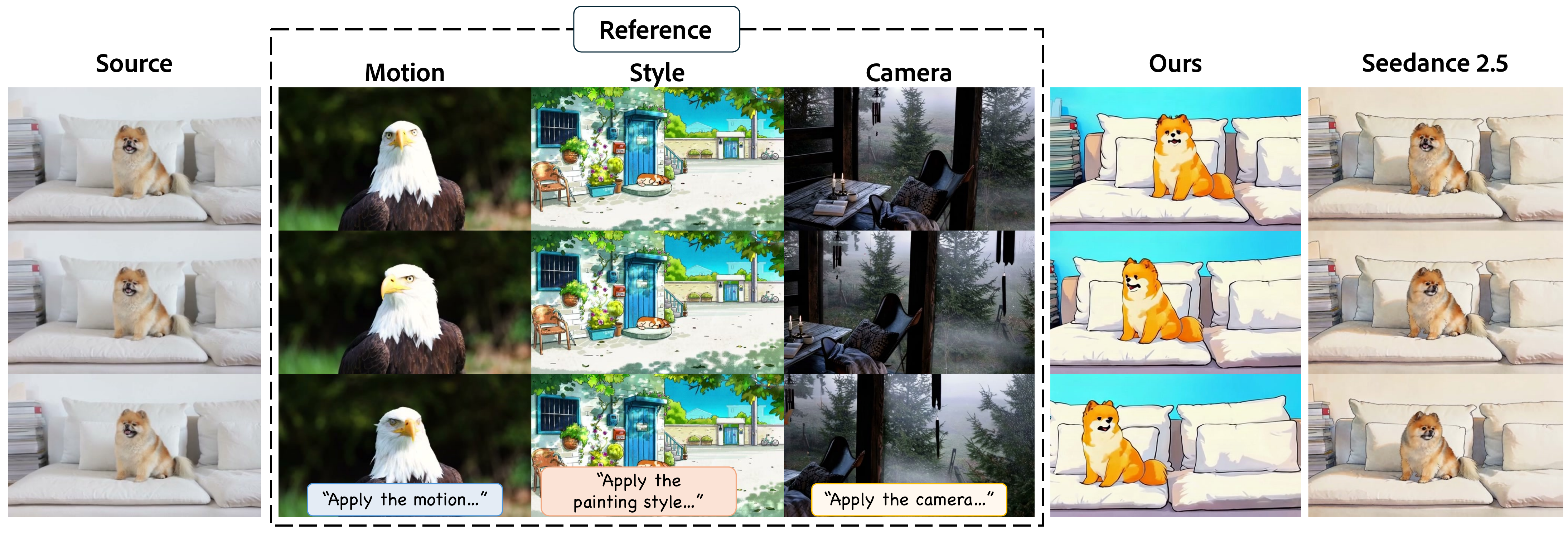}

\caption{Compositional generalization to three video references.}
\label{fig:compositional-generalization}
\end{figure}

\noindent\textbf{Generalization to more references.}
\label{sec:compositional-generalization}
Figure~\ref{fig:compositional-generalization} composes three video references specifying subject motion, camera trajectory, and style, exceeding the maximum of two visual references in any training sample.
\modelname{} follows all three references while preserving the source subject and scene thanks to the \rropername{} design, whereas Seedance~2.5 largely fails to follow the motion and camera references and only partially reproduces the requested style.

\FloatBarrier
\subsection{Inference Latency and Computational Cost}
\label{sec:inference-cost}
\begin{figure}[ht]
\centering
\vspace{-10pt}
\includegraphics[width=0.7\linewidth]{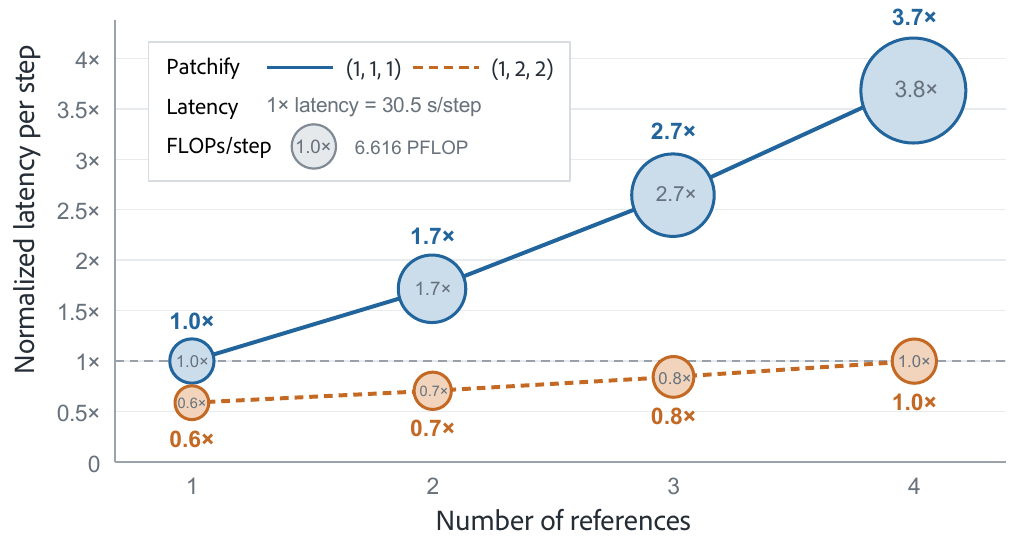}
\caption{Analysis of the inference latency and computational cost.}
\vspace{-10pt}
\label{fig:reference-scaling-cost}
\end{figure}

We measure DiT inference on a single NVIDIA A100 GPU, using 480P source and reference videos with 121 frames.
We vary both the number of references and the reference patch size in Mix-Patchify (Section~\ref{sec:mix-patchify}).
As Figure~\ref{fig:reference-scaling-cost} shows, more references and finer patches increase both latency and computational cost by lengthening the DiT input sequence.
Mix-Patchify reduces both costs, with a larger advantage as the number of references grows.

\FloatBarrier

\FloatBarrier
\section{Conclusion}
\label{sec:conclusion}

We introduced \frameworkname{}, a unified framework for multimodal multi-reference video editing, in which text, image, and video references are organized through a common reference-bank interface and composed in variable numbers to control distinct attributes of a source video.
\modelname{} integrates an MLLM with a single-stream DiT, employs Mix-Patchify to enable efficient conditioning on multiple video references, and adopts \rropername{} to generalize beyond the number of references observed during training.
To address the scarcity of video-reference editing data and benchmarks, we constructed \dataname{} through synthesis pipelines for Camera, Motion, Content, Style, and VFX, as well as compositions of two references, and established \benchname{} for the systematic evaluation of reference adherence, visual quality, and source preservation.
Extensive experiments demonstrate that \modelname{} outperforms both task-specific baselines and commercial systems on \benchname{}.
Further analyses verify that \modelname{} exploits the visual content of each reference, remains robust to the ordering of references, and generalizes to reference compositions beyond the training distribution.

\bibliography{references}
\bibliographystyle{assets/plainnat}

\clearpage
\appendix
\section{Interleaved Sequence Specification}
\label{app:conditioning-protocol}
\subsection{System Prompt}
\label{app:system-prompt}
The system prompt $T_{\mathrm{sys}}$ is shared by all reference configurations:
\begin{quote}
\small
You receive a source video, then one or more edit banks, each containing a task definition, an editing instruction, and optionally its own reference image or video. Apply each bank using only its own reference, take nothing else from any reference, and keep everything the banks do not change as in the source. Output the edited video.
\end{quote}

\subsection{Source and Output Markers}
\label{app:markers}
Two fixed text markers delimit the roles of the videos in the sequence.
The source marker $T_{\mathrm{src}}$, \texttt{This is the source video to edit:}, immediately precedes the source video and identifies it as the video whose unedited content must be preserved.
The output marker $T_{\mathrm{out}}$, \texttt{The output video is:}, immediately precedes the target tokens and signals where the edited video is generated.

\subsection{Reference Banks}
\label{app:reference-banks}
Each bank $B_i$ is a self-contained edit clause consisting of three parts.

\noindent\textbf{Task definition.}
The task definition $T_i^{\mathrm{task}}$ fixes the category of the bank and restricts what may be taken from its reference, while requiring the remaining source content to be preserved.
It thereby specifies \emph{which attribute} to extract, independently of the free-form text instruction.
Table~\ref{tab:task-definitions} lists the task definition of each category verbatim.

\begin{table}[ht]
\centering
\caption{Task definition $T_i^{\mathrm{task}}$ of each category}
\label{tab:task-definitions}
\begingroup
\small
\renewcommand{\arraystretch}{1.2}
\begin{tabular}{@{}lp{0.82\linewidth}@{}}
\toprule
Category & Task definition\\
\midrule
Camera & ``This is a camera edit --- change only the camera trajectory, keep everything else exactly as in the source.''\\
Motion & ``This is a motion edit --- change only the motion, keep everything else exactly as in the source.''\\
Content & ``This is a content edit --- change only the indicated subject or background, keep everything else exactly as in the source.''\\
VFX & ``This is a visual effect edit --- add only the indicated phenomenon, keep everything already in the source unchanged.''\\
Style & ``This is a style edit --- change only the lighting or stylization, keep everything else exactly as in the source.''\\
\bottomrule
\end{tabular}
\endgroup
\end{table}

\noindent\textbf{Editing instruction.}
The editing instruction $T_i^{\mathrm{edit}}$ is a free-form description of the concrete edit, for example, ``Replace the sky with the dark tornado storm cloud.''

\noindent\textbf{Visual reference.}
The optional visual reference $V_i$ is an image or a video that provides the content of the requested attribute.

For a text-only bank, $T_i^{\mathrm{task}}=\emptyset$ and $V_i=\emptyset$, so the bank contains only the editing instruction.
When an example contains several banks, they are concatenated after the source video, each with its own task definition, instruction, and reference.
The same layout therefore covers text-only editing (\editsetting{T}), single visual references (\editsetting{I} and \editsetting{V}), and their compositions (e.g., \editsetting{IV} and \editsetting{VV}) without any change to the model.

\section{Implementation Details}
\label{app:training-details}

\subsection{Training Objective}
\label{sec:training-objective}
Let $Z_0=\mathcal{E}_{\mathrm{VAE}}(V_{\mathrm{tgt}})$ be the clean target VAE latent.
At noise level $t\in[0,1]$, with $\epsilon\sim\mathcal{N}(0,I)$, we construct the noisy latent $Z_t=(1-t)Z_0+t\epsilon$.
The DiT processes the interleaved sequence $\mathcal{X}'$, in which $Z_t$ occupies the target positions, and predicts the velocity $u_\theta(Z_t,t)$ at these target positions. It is optimized with the flow-matching loss~\citep{lipman2022flowmatching}:
\begin{equation}
\mathcal{L}_{\mathrm{FM}}
=\mathbb{E}_{Z_0,t,\epsilon}\!
\left\|u_\theta(Z_t,t)-(Z_0-\epsilon)\right\|_2^2.
\label{eq:flow-matching-loss}
\end{equation}

\subsection{Mix-Patchify}
\label{app:mix-patchify}
The patchification layer converts continuous VAE latents into DiT tokens.
For a visual input $V$, the VAE encoder $\mathcal{E}_{\mathrm{VAE}}$ produces $Z=\mathcal{E}_{\mathrm{VAE}}(V)\in\mathbb{R}^{L\times H\times W\times C}$, where $L$, $H$, $W$, and $C$ denote the number of latent frames, latent height, latent width, and latent channel dimension, respectively.
A patch size $p=(p_t,p_h,p_w)$ specifies the number of latents grouped along the temporal, height, and width axes, respectively.
We partition $Z$ into $N_p$ non-overlapping patches, with zero-padding at the boundaries, where
\begin{equation}
N_p=\left\lceil\frac{L}{p_t}\right\rceil
\left\lceil\frac{H}{p_h}\right\rceil
\left\lceil\frac{W}{p_w}\right\rceil.
\label{eq:patch-token-count}
\end{equation}
For $j=1,\ldots,N_p$, let $\mathcal{P}_j$ denote the set of latent positions in the $j$-th patch, which contains $r=p_t \cdot p_h \cdot p_w$ latents.
The corresponding patch $Z_{\mathcal{P}_j}$ is flattened and mapped to a DiT token $\mathbf{e}_j$ by an MLP $\mathbb{P}_p$ specific to patch size $p$:
\begin{equation}
\mathbf{e}_j=\mathbb{P}_p(\operatorname{vec}(Z_{\mathcal{P}_j}))\in\mathbb{R}^{D},
\label{eq:mix-patchify}
\end{equation}
where $D$ is the DiT token dimension and $\operatorname{vec}(Z_{\mathcal{P}_j})\in\mathbb{R}^{rC}$ is the channel-wise concatenation of the $r$ latents in $Z_{\mathcal{P}_j}$.
Each visual input $V$ is thus represented by $N_p$ DiT tokens.
The pretrained DiT uses $p=(1,1,1)$ for fine-grained denoising.
We retain this granularity for the source and target tokens to preserve detailed source conditioning.
Reference tokens, however, do not all require such fine spatial granularity.
We therefore introduce \emph{Mix-Patchify} (MP), which represents each reference at a spatial granularity suited to its editing task while keeping $p_t=1$ throughout to preserve fine-grained temporal resolution for dynamic attributes.
For a patch size $p$, MP reduces the token count by up to a factor of $p_t \cdot p_h \cdot p_w$.
This substantially lowers training and inference costs, particularly with multiple references, while maintaining performance and facilitating convergence.

In practice, the task category of each reference bank determines its patch size: Content references retain $(1,1,1)$ patches to preserve identity and fine appearance details, while Camera, Motion, Style, and VFX references use spatially coarser $(1,2,2)$ patches.
To initialize a coarse-patch MLP $\mathbb{P}_p$ from the pretrained model, we set it to produce the average of the outputs of the pretrained fine-grained MLP $\mathbb{P}_{(1,1,1)}$ over the $r$ latents in each patch $Z_{\mathcal{P}_j}$, and then optimize it without this constraint.
To maintain a shared RoPE space across granularities, each coarse token $\mathbf{e}_j$ takes the $(y,x,\tau)$ coordinates of the top-left latent of its patch $\mathcal{P}_j$ on the original latent grid.

\section{Benchmark Evaluation Protocol}
\label{app:benchmark-evaluation}

\subsection{Category-Specific Evaluation}
\benchname{} evaluates single-video-reference editing across Camera, Motion, Content, Style, and VFX.
Video-reference editing raises three separate questions: whether the output adopts the requested dynamic attribute from the reference, whether the result looks and moves convincingly, and whether everything else from the source remains intact.
A model could produce an attractive video while ignoring the requested edit, or copy the desired attribute along with unrelated reference content; neither is a successful edit.
We use Gemini~3.8~Flash~\citep{google2026gemini38flash} as a VLM judge for three separate criteria: Reference Adherence (RA) assesses the requested reference attribute, Visual Quality (VQ) assesses the output quality, and Source Preservation (SP) assesses source information that should remain unchanged, including unwanted reference-content transfer.
The judge returns structured labels that we convert into scores deterministically; the complete category-specific prompts are listed in Appendix~\ref{app:judge-prompts}.

\subsection{Deterministic Scoring}
\label{app:deterministic-scoring}
The judge outputs only closed-set labels, which deterministic rules convert into scores on a $0$--$5$ scale (Table~\ref{tab:scoring-rules}).
Each criterion first receives a base score by averaging the scores of its labels; category-specific caps then penalize failures that the average alone would miss.
Leakage is graded by the number of source attributes that the SP judge labels as imported from the reference: two or more clear imports are \emph{major}, one clear import is \emph{minor}, and only possible imports are \emph{suspected}.
The composite Score is
\begin{equation}
\mathrm{Score}=0.45\,\mathrm{RA}+0.25\min(\mathrm{VQ},\mathrm{RA})+0.30\,\mathrm{SP},
\label{eq:composite-score}
\end{equation}
where bounding VQ by RA prevents visually appealing outputs that ignore the requested edit from scoring well, and the leakage caps prevent high scores on other criteria from compensating for severe unwanted changes.
All caps are applied per example before averaging; consequently, the reported mean RA, VQ, and SP do not in general determine the mean Score, and we report VQ before the $\min(\cdot,\mathrm{RA})$ bound so that visual quality can be distinguished from editing success.

\begin{table}[ht]
\centering
\caption{Conversion of judge labels into scores. The label scores of the judged items are averaged into a base score, and the listed caps are then applied.}
\label{tab:scoring-rules}
\begingroup
\footnotesize
\setlength{\tabcolsep}{4pt}
\renewcommand{\arraystretch}{1.2}
\begin{tabular}{@{}>{\raggedright\arraybackslash}p{0.13\linewidth}>{\raggedright\arraybackslash}p{0.21\linewidth}>{\raggedright\arraybackslash}p{0.25\linewidth}>{\raggedright\arraybackslash}p{0.33\linewidth}@{}}
\toprule
Criterion & Judged items & Label $\rightarrow$ score & Caps\\
\midrule
RA (Camera) & Direction, extent, timing, framing & match / partial / mismatch $\rightarrow 5/2.5/0$ & None (preservation and leakage are judged by SP)\\
\midrule
RA (Content, Style, VFX) & Category-specific properties & match / partial / mismatch $\rightarrow 5/2.5/0$ & Wrong localization $\rightarrow0$; spillover $\rightarrow\le2.5$; Style broken structure $\rightarrow\le1.25$; Content clear extra import $\rightarrow\le2.5$\\
\midrule
RA (Motion) & Pose match at six moments & match / partial / mismatch $\rightarrow 5/2.5/0$ & Reference share of appearance votes $\ge0.5\rightarrow0$, $\ge0.2\rightarrow\le2.5$; subject absent in more than half of the moments $\rightarrow0$\\
\midrule
VQ & Worst defect in appeal, plausibility, coherence & none / subtle / noticeable / obvious $\rightarrow 5/3.75/2/0$ & None\\
\midrule
SP & Source attributes & preserved / side effect / altered / absent / from reference $\rightarrow 5/5/2.5/1.25/0$ & Different video $\rightarrow0$; partly changed $\rightarrow\le4$; leakage: suspected $\le3.125$, minor $\le1.25$, major $0$\\
\midrule
Score & RA, VQ, SP & Eq.~(\ref{eq:composite-score}) & Major leakage $\rightarrow0$; minor leakage or $\mathrm{SP}\le1.25\rightarrow\le1.25$; suspected leakage $\rightarrow\le3.125$\\
\bottomrule
\end{tabular}
\endgroup
\end{table}

\subsection{Camera Trajectory Errors}
\label{app:camera-metrics}
For each Camera case, we sample $N=32$ temporally aligned frames from the reference and output videos and estimate their camera poses independently with VGGT-$\Omega$~\citep{wang2026vggtomega}.
Let $E_i^v$ be the estimated world-to-camera extrinsic matrix of frame $i$ in video $v\in\{\mathrm{ref},\mathrm{out}\}$, and let $T_i^v=(E_i^v)^{-1}$ be the corresponding camera-to-world pose.
We express each trajectory relative to its own first frame and normalize its translations by the maximum displacement from the first camera:
\begin{equation}
A_i^v=(T_0^v)^{-1}T_i^v=\begin{bmatrix}R_i^v&\mathbf{p}_i^v\\ \mathbf{0}^{\top}&1\end{bmatrix},
\qquad
\hat{\mathbf{p}}_i^v=\frac{\mathbf{p}_i^v}{\max_j\|\mathbf{p}_j^v\|_2+\epsilon}.
\label{eq:relative-pose}
\end{equation}
Expressing both trajectories in their own first-frame coordinates removes the arbitrary world frame of each reconstruction; no additional rigid alignment or temporal warping is applied between them.
Because both relative trajectories start at the identity, the first frame contributes no error; following CamI2V~\citep{zheng2024cami2v}, we sum the per-frame errors over the remaining $N-1=31$ frames:
\begin{equation}
\begin{aligned}
\mathrm{RotErr}&=\sum_{i=1}^{N-1}\arccos\!\left(\frac{\operatorname{tr}\big((R_i^{\mathrm{ref}})^{\top}R_i^{\mathrm{out}}\big)-1}{2}\right),\\
\mathrm{TransErr}&=\sum_{i=1}^{N-1}\big\|\hat{\mathbf{p}}_i^{\mathrm{ref}}-\hat{\mathbf{p}}_i^{\mathrm{out}}\big\|_2,\\
\mathrm{CamMC}&=\sum_{i=1}^{N-1}\big\|[R_i^{\mathrm{ref}}\mid\hat{\mathbf{p}}_i^{\mathrm{ref}}]-[R_i^{\mathrm{out}}\mid\hat{\mathbf{p}}_i^{\mathrm{out}}]\big\|_F,
\end{aligned}
\label{eq:camera-errors}
\end{equation}
where the argument of $\arccos$ is clipped to $[-1,1]$ and RotErr is measured in radians.
We then average each error over the Camera cases of \benchname{}.
Since each trajectory is normalized independently and camera intrinsics are not compared, these errors capture motion direction, trajectory shape, and timing, but not absolute translation magnitude or focal-length changes such as zooms.

\section{Instruction-Based Video Editing Benchmarks}
\label{app:additional-results}

Although \modelname{} focuses on reference-guided video editing, it should retain the foundational text-based editing capability on which such editing builds.
We therefore evaluate \editsetting{T} editing on VEFX-Bench~\citep{gao2026vefx} and ReCo-Bench~\citep{zhang2025reco}, following their official evaluation suites.
VEFX-Bench uses VEFX-Reward to score Instruction Following (IF), Rendering Quality (RQ), and Edit Exclusivity (EE) on a $1$--$4$ scale, together with their official geometric aggregate (GeoAgg).
ReCo-Bench reports editing accuracy, video naturalness, video quality, and Total, their arithmetic mean, separately for the Add, Remove, Replace, and Style tasks, evaluated by Gemini-2.5-Flash.

\begin{table}[ht]
\centering
\caption{VEFX-Bench leaderboard for \editsetting{T}.}
\label{tab:vefx-bench}
\begingroup
\newcommand{\commercialtext}[1]{\textcolor{black!45}{#1}}
\footnotesize
\setlength{\tabcolsep}{2pt}
\renewcommand{\arraystretch}{1.05}
\begin{tabular*}{0.6\linewidth}{@{}l@{\extracolsep{\fill}}cccc@{}}
\toprule
Model & IF $\uparrow$ & RQ $\uparrow$ & EE $\uparrow$ & GeoAgg $\uparrow$\\
\midrule\midrule
\multicolumn{5}{@{}c@{}}{\textcolor{black!65}{\textsc{Commercial}}}\\
\commercialtext{Kling o3 Omni} & \commercialtext{3.033} & \commercialtext{3.588} & \commercialtext{3.043} & \commercialtext{3.057}\\
\commercialtext{Kling o1} & \commercialtext{3.040} & \commercialtext{3.534} & \commercialtext{2.976} & \commercialtext{2.985}\\
\commercialtext{Runway Gen-4.5} & \commercialtext{2.817} & \commercialtext{3.319} & \commercialtext{2.923} & \commercialtext{2.912}\\
\commercialtext{Seedance 2.0} & \commercialtext{2.811} & \commercialtext{3.421} & \commercialtext{3.088} & \commercialtext{2.766}\\
\commercialtext{Grok Imagine} & \commercialtext{2.606} & \commercialtext{3.346} & \commercialtext{3.376} & \commercialtext{2.723}\\
\commercialtext{Wan 2.6} & \commercialtext{2.012} & \commercialtext{3.317} & \commercialtext{2.446} & \commercialtext{2.146}\\
\midrule
\multicolumn{5}{@{}c@{}}{\textcolor{black!65}{\textsc{Open-source}}}\\
UniVideo & 2.294 & 3.266 & \textbf{3.091} & 2.516\\
VACE & 2.027 & 3.172 & 1.180 & 1.775\\
\midrule
\textbf{\frameworkname{}} & \textbf{2.699} & \textbf{3.315} & 2.615 & \textbf{2.708}\\
\bottomrule
\end{tabular*}
\endgroup
\end{table}

We report \editsetting{T} results on VEFX-Bench~\citep{gao2026vefx} in Table~\ref{tab:vefx-bench}.
Overall, \modelname{} outperforms the open-source baselines~\citep{wei2025univideo,jiang2025vace}, as well as Wan~2.6~\citep{alibabacloud2025wan26}, and performs on par with commercial models such as Grok Imagine~\citep{xai2026grokimagine} and Seedance~2.0~\citep{bytedance2026seedance20}.
This indicates that, although \modelname{} focuses on reference-guided video editing, it retains the foundational text-based editing capability.
Table~\ref{tab:reco-bench} reports the ReCo-Bench leaderboard, and boldface marks the best value within each task and metric.
\begin{table}[!htb]
\centering
\caption{ReCo-Bench leaderboard for \editsetting{T} across four tasks: Add, Remove, Replace, Style.}
\label{tab:reco-bench}
\begingroup
\small
\setlength{\tabcolsep}{8pt}
\renewcommand{\arraystretch}{1.05}
\begin{tabular}{@{}lcccc@{}}
\toprule
Model & \shortstack{Editing\\accuracy $\uparrow$} & \shortstack{Video\\naturalness $\uparrow$} & \shortstack{Video\\quality $\uparrow$} & Total $\uparrow$\\
\midrule
\multicolumn{5}{@{}l}{\textit{Add}}\\
InsViE & 2.60 & 3.10 & 3.46 & 3.05\\
Lucy-Edit & 6.47 & 5.70 & 6.77 & 6.31\\
Ditto & 6.70 & 7.57 & 8.41 & 7.56\\
ReCo & 8.54 & 7.55 & 8.61 & 8.23\\
ReCo\_Ref & 7.96 & 7.39 & 8.69 & 8.02\\
VInO & 9.44 & 8.00 & 9.11 & 8.85\\
Mamoda2.5 & 9.31 & 8.27 & 8.83 & 8.80\\
\textbf{\frameworkname{}} & \textbf{9.55} & \textbf{8.56} & \textbf{9.12} & \textbf{9.08}\\
\midrule
\multicolumn{5}{@{}l}{\textit{Remove}}\\
InsViE & 2.44 & 3.76 & 3.29 & 3.16\\
VACE & 4.57 & 5.43 & 5.56 & 5.19\\
ReCo & 7.28 & 6.90 & 6.82 & 7.00\\
ReCo\_Ref & 8.06 & 7.45 & 7.44 & 7.65\\
VInO & 8.39 & 8.00 & 8.10 & 8.16\\
Mamoda2.5 & \textbf{9.15} & \textbf{8.50} & \textbf{8.35} & \textbf{8.67}\\
\textbf{\frameworkname{}} & 7.77 & 7.46 & 7.30 & 7.51\\
\midrule
\multicolumn{5}{@{}l}{\textit{Replace}}\\
InsViE & 2.10 & 3.91 & 3.49 & 3.17\\
Lucy-Edit & 7.08 & 6.21 & 6.88 & 6.72\\
Ditto & 4.56 & 7.21 & 7.96 & 6.58\\
ReCo & 9.43 & 8.01 & 8.77 & 8.74\\
ReCo\_Ref & 8.35 & 7.67 & 8.38 & 8.13\\
VInO & 9.17 & 8.18 & 8.74 & 8.70\\
Mamoda2.5 & \textbf{9.60} & 8.64 & 8.95 & \textbf{9.06}\\
\textbf{\frameworkname{}} & 9.10 & \textbf{8.70} & \textbf{8.96} & 8.92\\
\midrule
\multicolumn{5}{@{}l}{\textit{Style}}\\
InsViE & 8.17 & 8.21 & 7.35 & 7.91\\
Lucy-Edit & 4.65 & 4.67 & 5.17 & 4.83\\
Ditto & 9.20 & 9.07 & 8.77 & 9.01\\
ReCo & \textbf{9.42} & 9.19 & 8.90 & \textbf{9.17}\\
ReCo\_Ref & 8.81 & \textbf{9.27} & \textbf{9.29} & 9.12\\
VInO & 9.05 & 9.25 & 9.01 & 9.10\\
Mamoda2.5 & 8.95 & 9.19 & 8.83 & 8.99\\
\textbf{\frameworkname{}} & 8.37 & 8.80 & 8.39 & 8.52\\
\bottomrule
\end{tabular}
\endgroup
\end{table}

\FloatBarrier

\FloatBarrier
\Needspace{28\baselineskip}
\section{Limitations}
\label{sec:limitations}
\noindent
\begin{minipage}{\linewidth}
\begin{wrapfigure}{r}{0.5\linewidth}
\centering
\includegraphics[width=\linewidth]{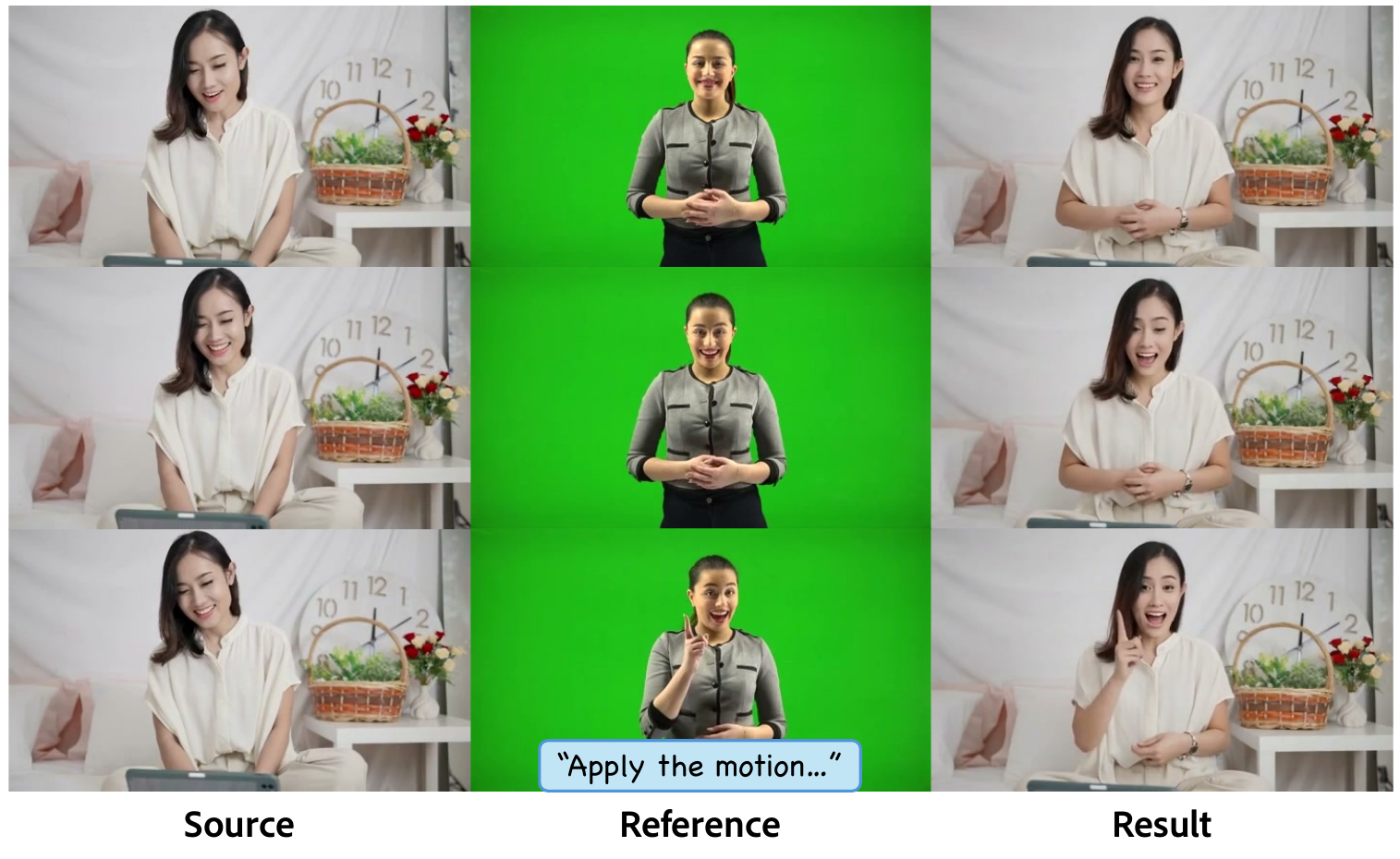}
\caption{Coupled transfer of facial expression and body pose from a Motion reference.}
\label{fig:limitation}
\end{wrapfigure}

\noindent
\modelname{} supports attribute-level control but cannot yet reliably disentangle finer-grained factors within an attribute.
For example, the model may transfer facial expressions and body poses together from a motion reference, making it difficult to apply either independently (Figure~\ref{fig:limitation}).
Scaling to larger reference compositions is also constrained by the difficulty of constructing high-quality training data in which the target video incorporates all requested edits.
Although our reserved-slot design supports generalization beyond the reference counts seen during training, performance degrades substantially with four or more references.
Our training data also underrepresent content beyond natural scenes, particularly VFX involving text rendering, indicating a need for greater diversity in visual content and editing operations.
Finally, training is currently limited to resolutions up to 480P, and the effectiveness of our framework for higher-resolution editing at 720P or 4K remains unexplored.
\par
\end{minipage}

\section{VLM Judge Prompts}
\label{app:judge-prompts}
This section lists the complete prompts that Gemini~3.8~Flash~\citep{google2026gemini38flash} receives when scoring \benchname{}.
Each category has its own Reference Adherence (RA) and Source Preservation (SP) prompts, which state explicitly what may change, what must be preserved, and which side effects of the edit are permitted; the Visual Quality (VQ) prompt is shared across categories.
Placeholders in braces, such as \verb|{instruction}| and \verb|{transferred_attribute}|, are filled from the annotation of each case, and \verb|[attach: ...]| marks where the visual evidence is inserted.
The judge never outputs numeric scores: it returns only closed-set labels and concrete, locatable evidence, from which the scores are computed deterministically as described in Appendix~\ref{app:benchmark-evaluation}.
The RA, VQ, and SP calls for a case are independent and share no context.
Visual evidence is provided as time-aligned frame grids.
\Needspace{20\baselineskip}
\subsection{Camera}
\label{app:prompts-camera}
The RA prompt compares only the camera motion of the reference and the output, using a two-row grid with eight time-aligned reference and output frames.
The equal-weight average of its four property labels gives the Camera RA.
Preservation of the source content and leakage of non-camera content from the reference are judged solely by the SP prompt.
  \begin{tcolorbox}[breakable, enhanced, title={Camera -- Reference Adherence (RA)},
    fonttitle=\bfseries\small, coltitle=white, colbacktitle=promptframe,
    colframe=promptframe, colback=promptback, boxrule=0.6pt, arc=1.5mm,
    left=5pt, right=5pt, top=4pt, bottom=4pt, before skip=8pt, after skip=10pt,
    fontupper=\fontsize{7.5}{9}\selectfont, parbox=false, label=prompt:camera-ra]
  \setlength{\parskip}{2pt}\raggedright
  \promptrole{System}
You judge whether output O follows the camera trajectory of reference R. R and O show different scenes by design. Compare camera movement only, not subjects, scene content, colour, or absolute object positions. Use background movement and parallax where visible; moving subjects alone are not evidence of camera movement.

Return one JSON object and nothing else. Do not output a numeric score.

\promptrole{User}
\begin{promptitems}
\item[\textit{instruction}] ``\{instruction\}''
\end{promptitems}

\promptattach{[attach: RO\_rows~--- one image, 2 rows x 8 columns.\\ Top = R (reference), bottom = O (output).\\ Read each row from left to right.]}

\prompthead{Step 1} Describe R's camera path and O's camera path concretely:
direction, amount of viewpoint change, when the move starts/turns/stops, and how the view changes. $\rightarrow$ ``R\_reading'', ``O\_reading''

\prompthead{Step 2} Compare O's camera movement against R's on four properties:
\begin{promptitems}
\item[\textit{direction}] dominant direction and movement type (e.g. pan, tilt, dolly, orbit, zoom); do not confuse subject motion with camera motion
\item[\textit{extent}] comparable relative amount of viewpoint change; do not demand equal pixel displacement in different scenes
\item[\textit{timing}] comparable onset, pauses, turns and speed progression over normalized time; do not penalize different clip durations
\item[\textit{framing}] comparable change in viewpoint or shot scale from start to end; do not compare absolute scene composition
\end{promptitems}

For each property use ``match'', ``partial'', or ``mismatch''. Judge only what the frames support; do not invent an exact 3D trajectory.

\promptreturn
\begin{lstlisting}[style=promptjson]
{"R_reading":"","O_reading":"",
 "per_property":{"direction":"match|partial|mismatch",
                 "extent":"match|partial|mismatch",
                 "timing":"match|partial|mismatch",
                 "framing":"match|partial|mismatch"},
 "reasoning":""}
\end{lstlisting}

  \end{tcolorbox}
  \begin{tcolorbox}[breakable, enhanced, title={Camera -- Source Preservation (SP)},
    fonttitle=\bfseries\small, coltitle=white, colbacktitle=promptframe,
    colframe=promptframe, colback=promptback, boxrule=0.6pt, arc=1.5mm,
    left=5pt, right=5pt, top=4pt, bottom=4pt, before skip=8pt, after skip=10pt,
    fontupper=\fontsize{7.5}{9}\selectfont, parbox=false, label=prompt:camera-sp]
  \setlength{\parskip}{2pt}\raggedright
  \promptrole{System}
You judge preservation for camera-trajectory transfer. O should retain S's scene and content while its camera trajectory changes. Check whether S's non-camera content survives and whether non-camera content from R was copied into O.

Changes in viewpoint, crop, perspective, parallax, scale and visibility caused by the new camera are allowed. The requested camera movement from R is not reference-content leakage.

Return one JSON object and nothing else. Do not output a numeric score.

\promptrole{User}
\begin{promptitems}
\item[\textit{instruction}] ``\{instruction\}''
\end{promptitems}

\promptattach{[attach: SOR\_rows~--- one image, 3 rows x 6 columns.\\ Top = S (source), middle = O (output), bottom = R (reference).\\ Columns correspond to the same normalized moments.\\ Compare O with S for preservation; use R to identify copied content.]}

\prompthead{Part A} Ignoring the allowed camera change, does O still depict S's
video? Choose ``preserved'', ``partly\_changed'', or ``different\_video''. Give one concrete reason. $\rightarrow$ ``preservation\_overall'', ``overall\_reason''

\prompthead{Part B} List the checkable attributes of S using these names:
\begin{promptitems}
\item[\textit{setting}] place and environment
\item[\textit{subjects}] people/animals and their identity and appearance
\item[\textit{objects}] notable props and objects
\item[\textit{lighting}] lighting direction and character
\item[\textit{style}] palette and visual treatment
\item[\textit{action}] what the subjects do
\end{promptitems}

For each, describe its version in S and O, then choose:
\begin{promptitems}
\item[\textit{``preserved''}] O keeps S's version
\item[\textit{``side\_effect''}] changed only because of the new viewpoint, crop or occlusion
\item[\textit{``altered''}] changed into something else
\item[\textit{``from\_reference''}] distinctive non-camera content from R appeared in O
\item[\textit{``absent''}] gone despite still being checkable from O's new viewpoint
\end{promptitems}

Use ``clear'' or ``possible'' for certainty. Skip an attribute if the new viewpoint makes it impossible to check. Do not list camera or framing: they are the intended edit.

\promptreturn
\begin{lstlisting}[style=promptjson]
{"preservation_overall":"preserved|partly_changed|different_video",
 "overall_reason":"",
 "attributes":[{"name":"setting|subjects|objects|lighting|style|action",
                "in_source":"","in_output":"",
                "verdict":"preserved|side_effect|altered|from_reference|absent",
                "certainty":"clear|possible"}]}
\end{lstlisting}

  \end{tcolorbox}

\Needspace{20\baselineskip}
\subsection{Motion}
\label{app:prompts-motion}
For each of six time-aligned moments, the RA judge receives one image with the source, reference, and output frames side by side (S\,$|$\,R\,$|$\,O) in a separate call.
The six pose labels are averaged, and the 18 appearance votes (three attributes over six moments) determine whether the subject's identity was copied from the reference.
  \begin{tcolorbox}[breakable, enhanced, title={Motion -- Reference Adherence (RA), one call per sampled moment},
    fonttitle=\bfseries\small, coltitle=white, colbacktitle=promptframe,
    colframe=promptframe, colback=promptback, boxrule=0.6pt, arc=1.5mm,
    left=5pt, right=5pt, top=4pt, bottom=4pt, before skip=8pt, after skip=10pt,
    fontupper=\fontsize{7.5}{9}\selectfont, parbox=false, label=prompt:motion-ra]
  \setlength{\parskip}{2pt}\raggedright
  \promptrole{System}
You are shown one moment from three videos, side by side at the same normalized time:
\begin{promptitems}
\item[\textit{S}] source (the person whose look must be kept)
\item[\textit{R}] reference (the person whose motion must be copied)
\item[\textit{O}] output (the edited result)
\end{promptitems}
S and R show different people and different places by design.

Judge this single moment only. Two things are being judged and they must never be confused:
\begin{promptitems}
\item[\textit{Pose}] the body configuration at this instant. O should match R.
\item[\textit{Identity}] what the subject looks like. O should match S, not R.
\end{promptitems}
A very close pose match is a success, never evidence that identity was copied.

Return one JSON object and nothing else. Do not output any numeric score.

\promptrole{User}
\begin{promptitems}
\item[\textit{instruction}] ``\{instruction\}''
\item[\textit{motion}] ``\{transferred\_attribute\}''
\end{promptitems}
\promptattach{[attach: one image, 1 row x 3 columns -- S \textbar{} R \textbar{} O at the same moment]}

\prompthead{Step 1} Describe R's body configuration at this instant: torso orientation, where each limb is, head direction, and what the action looks like frozen here. $\rightarrow$ ``R\_pose''
\prompthead{Step 2} Describe O's body configuration the same way. $\rightarrow$ ``O\_pose''
\prompthead{Step 3} ``pose\_match'': how well does O's pose reproduce R's pose at this moment?
\begin{promptitems}
\item[\textit{``match''}] the same configuration; limbs in the same places
\item[\textit{``partial''}] the same general action but limbs or orientation clearly differ
\item[\textit{``mismatch''}] a different configuration, or O's subject is doing something else
\end{promptitems}
Judge the body only. A different person, different clothes or a different room must not lower this label.
\prompthead{Step 4} Now ignore all pose. Looking only at how the subject looks, say for each attribute whether O's subject resembles S's or R's:
\begin{promptitems}
\item[\textit{clothing}] garment type and colour
\item[\textit{hair}] colour, length, style
\item[\textit{build}] body type and proportions
\end{promptitems}
Use ``unclear'' only when the subject is too small or occluded to tell.
\prompthead{Step 5} ``subject\_present'': is there a recognisable subject performing an action in O at all?
\begin{promptitems}
\item[\textit{``yes''}] \textbar{} ``no''
\end{promptitems}

\promptreturn
\begin{lstlisting}[style=promptjson]
{"R_pose":"","O_pose":"",
 "pose_match":"match|partial|mismatch",
 "appearance":{"clothing":"source|reference|unclear","hair":"source|reference|unclear",
               "build":"source|reference|unclear"},
 "subject_present":"yes|no",
 "reasoning":""}
\end{lstlisting}

  \end{tcolorbox}
  \begin{tcolorbox}[breakable, enhanced, title={Motion -- Source Preservation (SP)},
    fonttitle=\bfseries\small, coltitle=white, colbacktitle=promptframe,
    colframe=promptframe, colback=promptback, boxrule=0.6pt, arc=1.5mm,
    left=5pt, right=5pt, top=4pt, bottom=4pt, before skip=8pt, after skip=10pt,
    fontupper=\fontsize{7.5}{9}\selectfont, parbox=false, label=prompt:motion-sp]
  \setlength{\parskip}{2pt}\raggedright
  \promptrole{System}
You judge preservation for motion transfer: whether an edited video kept everything it was supposed to keep.

The requested edit is:
\begin{promptitems}
\item[\textit{instruction}] ``\{instruction\}''
\item[\textit{transferred attribute}] ``\{transferred\_attribute\}''
\end{promptitems}

O should be S's video with only the transferred attribute changed. Two kinds of failure matter:
\begin{promptitems}
\item[\textit{(a)}] something of S's was lost or altered when it should have stayed;
\item[\textit{(b)}] something of R's, other than the transferred attribute, turned up in O.
\end{promptitems}

The transferred attribute appearing in O is correct, however strongly it resembles R's version. The subject's pose and limb positions will change~--- that is the requested edit and is never a preservation failure. The subject's identity and appearance changing is a failure, and is the most important thing you are checking.

Judge from the pixels. Return one JSON object and nothing else. Do not output any numeric score.

\promptrole{User}
\promptattach{[attach: SOR\_rows] (one image, 3 rows x 6 columns, columns are time-aligned:\\ Top row    = S, the source, at 6 timestamps, earliest on the left\\ Middle row = O, the output, at the same 6 timestamps\\ Bottom row = R, the reference, at the same 6 timestamps\\ Read along a row to see how that video evolves over time.\\ Compare O with S by looking one row up, and O with R one row down.)}

\prompthead{Part A} one overall judgement for the whole clip. Compare the middle row against the top row. Mentally set aside everything the edit was allowed to change. Ignoring that completely, does what remains still look like the same video?
\begin{promptitems}
\item[\textit{``preservation\_overall''}] ``preserved'' \textbar{} ``partly\_changed'' \textbar{} ``different\_video''
\item[\textit{``overall\_reason''}] one sentence saying what drove that answer.
\end{promptitems}

\prompthead{Part B} attribute-by-attribute comparison.
\prompthead{Step B1} Read the top row (S) and list the key attributes S actually has. Use these slots, skipping any that do not apply to this clip:
\begin{promptitems}
\item[\textit{identity}] the source subject's face, hair, clothing and body type
\item[\textit{setting}] where it takes place: background, terrain, architecture, interior
\item[\textit{objects}] notable props and objects
\item[\textit{layout}] spatial arrangement, composition, framing
\item[\textit{camera}] camera position and how it moves along the row
\item[\textit{lighting}] lighting direction, tone, time of day
\end{promptitems}
Describe S's version of each in a few concrete words $\rightarrow$ ``in\_source''

\prompthead{Step B2} For that same attribute, compare the middle row (O) against the top row (S) and against the bottom row (R), then choose one verdict:
\begin{promptitems}
\item[\textit{``preserved''}] O keeps S's version of it
\item[\textit{``side\_effect''}] changed only as an unavoidable consequence of the transferred attribute (a change forced by the new pose, such as a limb now occluding something)
\item[\textit{``altered''}] changed into something that is neither S's nor R's
\item[\textit{``from\_reference''}] O's version of it now matches R's instead of S's
\item[\textit{``absent''}] it is gone from O
\end{promptitems}
State your certainty~--- ``clear'' \textbar{} ``possible''~--- and describe what you actually see in O in a few concrete words $\rightarrow$ ``in\_output''

Do not list the motion, pose or limb positions~--- they are the transferred attribute. Judge `identity' strictly: if O's subject wears R's clothes, has R's hair or R's build, that is `from\_reference', not `altered'.

\promptreturn
\begin{lstlisting}[style=promptjson]
{"preservation_overall":"preserved|partly_changed|different_video",
 "overall_reason":"",
 "attributes":[{"name":"identity|setting|objects|layout|camera|lighting",
                "in_source":"","in_output":"",
                "verdict":"preserved|side_effect|altered|from_reference|absent",
                "certainty":"clear|possible"}]}
\end{lstlisting}

  \end{tcolorbox}

\Needspace{20\baselineskip}
\subsection{Content}
\label{app:prompts-content}
The RA prompt receives six-frame filmstrips of the reference and the output, together with the location of the requested element in each.
  \begin{tcolorbox}[breakable, enhanced, title={Content -- Reference Adherence (RA)},
    fonttitle=\bfseries\small, coltitle=white, colbacktitle=promptframe,
    colframe=promptframe, colback=promptback, boxrule=0.6pt, arc=1.5mm,
    left=5pt, right=5pt, top=4pt, bottom=4pt, before skip=8pt, after skip=10pt,
    fontupper=\fontsize{7.5}{9}\selectfont, parbox=false, label=prompt:content-ra]
  \setlength{\parskip}{2pt}\raggedright
  \promptrole{System}
You compare a reference clip R and an output clip O to decide whether one specific content element from R was correctly brought into O. That element may be a background, an object, a subject, or a visual attribute~--- the instruction says which. Compare only that element; never compare the rest of R's scene.

Bringing it in is the point of the edit. However strongly O resembles R on that one element, that is success, never evidence of copying.

Return one JSON object and nothing else. Do not output any numeric score~--- scores are computed downstream from your labels.

\promptrole{User}
\begin{promptitems}
\item[\textit{instruction}] ``\{instruction\}''
\item[\textit{element}] ``\{attribute\}''        (the content element that was requested)
\item[\textit{locus\_R}] ``\{ref\_locus\}''        (in R, where the element is)
\item[\textit{locus\_O}] ``\{edit\_locus\}''       (in O, where it should appear)
\end{promptitems}
\promptattach{[attach: R\_filmstrip] (6 frames)}
\promptattach{[attach: O\_filmstrip] (6 frames)}

\prompthead{Step 1} Read the instruction and choose exactly one edit type:
\begin{promptitems}
\item[\textit{``replace\_background''}] R's surroundings take over S's surroundings
\item[\textit{``replace\_object''}] one thing in the source is swapped for a thing from R
\item[\textit{``add\_object''}] a thing from R is added; nothing existing is removed
\item[\textit{``apply\_attribute''}] a property from R (pattern, colour, material) is applied to something that stays
\end{promptitems}
$\rightarrow$ ``edit\_type''

\prompthead{Step 2} From R at locus\_R, describe the element: what it is, its texture, colour, material and distinguishing details. $\rightarrow$ ``R\_reading''
\prompthead{Step 3} From O, state whether the element is present and describe it the same way. $\rightarrow$ ``O\_reading''

\prompthead{Step 4} For each property decide whether O matches R:
\begin{promptitems}
\item[\textit{identity}] it is recognisably the same thing
\item[\textit{appearance}] texture, colour, material detail match
\item[\textit{extent}] does it occupy the right scope in O? Judge by edit\_type: replace\_background $\rightarrow$ it fills the surroundings, and O's foreground subjects still sit in front of it, not erased
\begin{promptitems}
\item[\textit{replace\_object}] $\rightarrow$ right position, scale and orientation, standing where the thing it replaced used to be
\item[\textit{add\_object}] $\rightarrow$ right position, scale and orientation for O's scene
\item[\textit{apply\_attribute}] $\rightarrow$ applied to the thing named in the instruction, and to nothing else
\end{promptitems}
\item[\textit{integration}] lighting, shadow, contact and occlusion are consistent with O's scene
\end{promptitems}
A different underlying scene in R is expected and must never lower these labels.

\prompthead{Step 5} Set ``extra\_import'': besides the requested element, did anything else from R appear in O (R's other objects, R's subjects, R's surroundings, R's camera behaviour)? $\rightarrow$ ``none'' \textbar{} ``possible'' \textbar{} ``clear''

\prompthead{Step 6} Set ``localization'':
\begin{promptitems}
\item[\textit{``correct''}] only the intended region changed
\item[\textit{``leaks''}] the change spilled into regions it should not touch
\item[\textit{``wrong''}] the element is absent, or landed somewhere else entirely
\end{promptitems}

\promptreturn
\begin{lstlisting}[style=promptjson]
{"edit_type":"replace_background|replace_object|add_object|apply_attribute",
 "R_reading":"","O_reading":"",
 "per_property":{"identity":"match|partial|mismatch","appearance":"match|partial|mismatch",
                 "extent":"match|partial|mismatch","integration":"match|partial|mismatch"},
 "localization":"correct|leaks|wrong",
 "extra_import":"none|possible|clear",
 "reasoning":""}
\end{lstlisting}

  \end{tcolorbox}
  \begin{tcolorbox}[breakable, enhanced, title={Content -- Source Preservation (SP)},
    fonttitle=\bfseries\small, coltitle=white, colbacktitle=promptframe,
    colframe=promptframe, colback=promptback, boxrule=0.6pt, arc=1.5mm,
    left=5pt, right=5pt, top=4pt, bottom=4pt, before skip=8pt, after skip=10pt,
    fontupper=\fontsize{7.5}{9}\selectfont, parbox=false, label=prompt:content-sp]
  \setlength{\parskip}{2pt}\raggedright
  \promptrole{System}
You judge preservation for content transfer: whether an edited video kept everything it was supposed to keep.

The requested edit is:
\begin{promptitems}
\item[\textit{instruction}] ``\{instruction\}''
\item[\textit{transferred attribute}] ``\{transferred\_attribute\}''
\end{promptitems}

O should be S's video with only the transferred attribute changed. Two kinds of failure matter:
\begin{promptitems}
\item[\textit{(a)}] something of S's was lost or altered when it should have stayed;
\item[\textit{(b)}] something of R's, other than the transferred attribute, turned up in O.
\end{promptitems}

The transferred attribute appearing in O is correct, however strongly it resembles R's version. This task deliberately imports one element from R: the requested element above. That element arriving from R is correct and must never be reported as `from\_reference'. Anything else arriving from R is a failure.

Judge from the pixels. Return one JSON object and nothing else. Do not output any numeric score.

\promptrole{User}
\promptattach{[attach: SOR\_rows] (one image, 3 rows x 6 columns, columns are time-aligned:\\ Top row    = S, the source, at 6 timestamps, earliest on the left\\ Middle row = O, the output, at the same 6 timestamps\\ Bottom row = R, the reference, at the same 6 timestamps\\ Read along a row to see how that video evolves over time.\\ Compare O with S by looking one row up, and O with R one row down.)}

\prompthead{Part A} one overall judgement for the whole clip. Compare the middle row against the top row. Mentally set aside everything the edit was allowed to change. Ignoring that completely, does what remains still look like the same video?
\begin{promptitems}
\item[\textit{``preservation\_overall''}] ``preserved'' \textbar{} ``partly\_changed'' \textbar{} ``different\_video''
\item[\textit{``overall\_reason''}] one sentence saying what drove that answer.
\end{promptitems}

\prompthead{Part B} attribute-by-attribute comparison.
\prompthead{Step B1} Read the top row (S) and list the key attributes S actually has. Use these slots, skipping any that do not apply to this clip:
\begin{promptitems}
\item[\textit{setting}] where it takes place: background, terrain, architecture, interior
\item[\textit{subjects}] the people / animals / main entities present, and their identity
\item[\textit{objects}] notable props and objects other than the requested element
\item[\textit{layout}] spatial arrangement, composition, framing
\item[\textit{camera}] camera position and how it moves along the row
\item[\textit{lighting}] lighting direction, tone, time of day
\item[\textit{action}] what happens across the six columns
\end{promptitems}
Describe S's version of each in a few concrete words $\rightarrow$ ``in\_source''

\prompthead{Step B2} For that same attribute, compare the middle row (O) against the top row (S) and against the bottom row (R), then choose one verdict:
\begin{promptitems}
\item[\textit{``preserved''}] O keeps S's version of it
\item[\textit{``side\_effect''}] changed only as an unavoidable consequence of the transferred attribute (the requested element occluding or casting light on something)
\item[\textit{``altered''}] changed into something that is neither S's nor R's
\item[\textit{``from\_reference''}] O's version of it now matches R's instead of S's
\item[\textit{``absent''}] it is gone from O
\end{promptitems}
State your certainty~--- ``clear'' \textbar{} ``possible''~--- and describe what you actually see in O in a few concrete words $\rightarrow$ ``in\_output''

Do not list the requested element itself~--- its arrival from R is the point of the edit.

\promptreturn
\begin{lstlisting}[style=promptjson]
{"preservation_overall":"preserved|partly_changed|different_video",
 "overall_reason":"",
 "attributes":[{"name":"setting|subjects|objects|layout|camera|lighting|action",
                "in_source":"","in_output":"",
                "verdict":"preserved|side_effect|altered|from_reference|absent",
                "certainty":"clear|possible"}]}
\end{lstlisting}

  \end{tcolorbox}

\Needspace{20\baselineskip}
\subsection{Style}
\label{app:prompts-style}
The RA prompt receives six rows of time-aligned reference and output frame pairs, and additionally checks whether the stylization destroys the structure of the output.
  \begin{tcolorbox}[breakable, enhanced, title={Style -- Reference Adherence (RA)},
    fonttitle=\bfseries\small, coltitle=white, colbacktitle=promptframe,
    colframe=promptframe, colback=promptback, boxrule=0.6pt, arc=1.5mm,
    left=5pt, right=5pt, top=4pt, bottom=4pt, before skip=8pt, after skip=10pt,
    fontupper=\fontsize{7.5}{9}\selectfont, parbox=false, label=prompt:style-ra]
  \setlength{\parskip}{2pt}\raggedright
  \promptrole{System}
You compare two clips, R (reference) and O (output). R carries a specific lighting/colour style, and O was supposed to receive that same style. Decide how faithfully O reproduces it. The two clips show different scenes by design~--- compare only the style itself, never the surrounding content, and never penalise O for having a different scene than R. Return one JSON object and nothing else. Do not output any numeric score.

\promptrole{User}
\begin{promptitems}
\item[\textit{instruction}] ``\{instruction\}''
\item[\textit{style}] ``\{transferred\_attribute\}''
\end{promptitems}
scope:        whole frame
\promptattach{[attach: RO\_pairs] (6 rows; each row is R[t] on the left, O[t] on the right,\\ At the same normalized time)}

\prompthead{Step 1} From R, describe the style concretely on each property~--- palette, lighting, texture. $\rightarrow$ ``R\_reading''
\prompthead{Step 2} In O, describe the same three properties. $\rightarrow$ ``O\_reading''
\prompthead{Step 3} For each property decide whether O matches R. Judge the style only; a different underlying scene is expected and must not lower these labels.
\prompthead{Step 4} ``structure\_check'': judging O alone, are its objects and surfaces structurally coherent (``intact''), mildly deformed or smeared (``soft''), or clearly distorted, melted or unrecognisable (``broken'')? Over-stylisation that destroys structure must be caught
\begin{promptitems}
\item[\textit{here}] a beautiful grade over ruined geometry is not a success.
\end{promptitems}

\promptreturn
\begin{lstlisting}[style=promptjson]
{"R_reading":"","O_reading":"",
 "per_property":{"palette":"match|partial|mismatch","lighting":"match|partial|mismatch",
                 "texture":"match|partial|mismatch"},
 "structure_check":"intact|soft|broken",
 "reasoning":""}
\end{lstlisting}

  \end{tcolorbox}
  \begin{tcolorbox}[breakable, enhanced, title={Style -- Source Preservation (SP)},
    fonttitle=\bfseries\small, coltitle=white, colbacktitle=promptframe,
    colframe=promptframe, colback=promptback, boxrule=0.6pt, arc=1.5mm,
    left=5pt, right=5pt, top=4pt, bottom=4pt, before skip=8pt, after skip=10pt,
    fontupper=\fontsize{7.5}{9}\selectfont, parbox=false, label=prompt:style-sp]
  \setlength{\parskip}{2pt}\raggedright
  \promptrole{System}
You judge preservation for style transfer: whether an edited video kept everything it was supposed to keep.

The requested edit is:
\begin{promptitems}
\item[\textit{instruction}] ``\{instruction\}''
\item[\textit{transferred attribute}] ``\{transferred\_attribute\}''
\end{promptitems}

O should be S's video with only the transferred attribute changed. Two kinds of failure matter:
\begin{promptitems}
\item[\textit{(a)}] something of S's was lost or altered when it should have stayed;
\item[\textit{(b)}] something of R's, other than the transferred attribute, turned up in O.
\end{promptitems}

The transferred attribute appearing in O is correct, however strongly it resembles R's version. Palette, colour grading, contrast, lighting tone and surface texture are the transferred attribute~--- they are supposed to change, and changing them is never a failure. What must survive is structure: the objects, their shapes and positions, and the place itself.

Judge from the pixels. Return one JSON object and nothing else. Do not output any numeric score.

\promptrole{User}
\promptattach{[attach: SOR\_rows] (one image, 3 rows x 6 columns, columns are time-aligned:\\ Top row    = S, the source, at 6 timestamps, earliest on the left\\ Middle row = O, the output, at the same 6 timestamps\\ Bottom row = R, the reference, at the same 6 timestamps\\ Read along a row to see how that video evolves over time.\\ Compare O with S by looking one row up, and O with R one row down.)}

\prompthead{Part A} one overall judgement for the whole clip. Compare the middle row against the top row. Mentally set aside everything the edit was allowed to change. Ignoring that completely, does what remains still look like the same video?
\begin{promptitems}
\item[\textit{``preservation\_overall''}] ``preserved'' \textbar{} ``partly\_changed'' \textbar{} ``different\_video''
\item[\textit{``overall\_reason''}] one sentence saying what drove that answer.
\end{promptitems}

\prompthead{Part B} attribute-by-attribute comparison.
\prompthead{Step B1} Read the top row (S) and list the key attributes S actually has. Use these slots, skipping any that do not apply to this clip:
\begin{promptitems}
\item[\textit{setting}] where it takes place: background, terrain, architecture, interior
\item[\textit{subjects}] the people / animals / main entities present, and their identity
\item[\textit{objects}] notable props and objects
\item[\textit{layout}] spatial arrangement, composition, framing
\item[\textit{camera}] camera position and how it moves along the row
\item[\textit{action}] what happens across the six columns
\end{promptitems}
Describe S's version of each in a few concrete words $\rightarrow$ ``in\_source''

\prompthead{Step B2} For that same attribute, compare the middle row (O) against the top row (S) and against the bottom row (R), then choose one verdict:
\begin{promptitems}
\item[\textit{``preserved''}] O keeps S's version of it
\item[\textit{``side\_effect''}] changed only as an unavoidable consequence of the transferred attribute (a colour or tonal shift that the new style necessarily implies)
\item[\textit{``altered''}] changed into something that is neither S's nor R's
\item[\textit{``from\_reference''}] O's version of it now matches R's instead of S's
\item[\textit{``absent''}] it is gone from O
\end{promptitems}
State your certainty~--- ``clear'' \textbar{} ``possible''~--- and describe what you actually see in O in a few concrete words $\rightarrow$ ``in\_output''

Do not list lighting, colour, palette or texture~--- they are the transferred attribute. A useful test: ignore all colour and lighting. Does the remaining structure still look like S's?

\promptreturn
\begin{lstlisting}[style=promptjson]
{"preservation_overall":"preserved|partly_changed|different_video",
 "overall_reason":"",
 "attributes":[{"name":"setting|subjects|objects|layout|camera|action",
                "in_source":"","in_output":"",
                "verdict":"preserved|side_effect|altered|from_reference|absent",
                "certainty":"clear|possible"}]}
\end{lstlisting}

  \end{tcolorbox}

\Needspace{20\baselineskip}
\subsection{VFX}
\label{app:prompts-vfx}
The RA prompt receives a two-row grid with six time-aligned reference and output frames, and additionally labels where the effect appears and how long it remains visible.
  \begin{tcolorbox}[breakable, enhanced, title={VFX -- Reference Adherence (RA)},
    fonttitle=\bfseries\small, coltitle=white, colbacktitle=promptframe,
    colframe=promptframe, colback=promptback, boxrule=0.6pt, arc=1.5mm,
    left=5pt, right=5pt, top=4pt, bottom=4pt, before skip=8pt, after skip=10pt,
    fontupper=\fontsize{7.5}{9}\selectfont, parbox=false, label=prompt:vfx-ra]
  \setlength{\parskip}{2pt}\raggedright
  \promptrole{System}
You compare two clips, R (reference) and O (output). R exhibits a specific visual effect, and O was supposed to receive that same effect. Decide how faithfully O reproduces it. The two clips show different scenes by design~--- compare only the effect itself, never the surrounding content, and never penalise O for having a different scene than R. Return one JSON object and nothing else. Every field is a label chosen from the listed options or free text~--- never a number, never a rating.

\promptrole{User}
\begin{promptitems}
\item[\textit{instruction}] ``\{instruction\}''
\item[\textit{effect}] ``\{transferred\_attribute\}''
\end{promptitems}
target\_region:"\{edit\_locus\}"     (the region of O where the effect belongs)
\promptattach{[attach: RO\_strips] (one image, 2 rows x 6 columns.\\ Top row    = R at 6 timestamps, earliest on the left\\ Bottom row = O at the same 6 timestamps\\ Read each row left-to-right to see how the effect evolves;\\ Compare top and bottom within a column for the same moment.)}

\prompthead{Step 1} Reading the top row left to right, describe R's effect concretely on each property~--- appearance, colour, dynamics, integration. $\rightarrow$ ``R\_reading''
\prompthead{Step 2} Reading the bottom row the same way, state whether the effect is present in O and, if so, describe it on the same four properties. $\rightarrow$ ``O\_reading''
\prompthead{Step 3} For each property decide whether O matches R. Judge the effect only; a different underlying scene is expected and must not lower these labels.
\begin{promptitems}
\item[\textit{appearance}] the form, shape and material of the effect
\item[\textit{color}] its colour and colour temperature
\item[\textit{dynamics}] how it evolves across the row: onset, spread, decay, speed
\item[\textit{integration}] how it sits in the scene: lighting it casts, shadows, occlusion
\end{promptitems}
\prompthead{Step 4} ``localization'': is the effect in O in the target\_region (``correct''), somewhere else (``wrong''), or does it also bleed into regions it should not touch (``leaks'')? If the effect is entirely absent, use ``wrong''.
\prompthead{Step 5} ``effect\_duration'': across the 6 bottom-row frames, how long is the effect clearly visible?
\begin{promptitems}
\item[\textit{``absent''}] not visible in any frame
\item[\textit{``brief''}] visible in only one or two frames
\item[\textit{``intermittent''}] comes and goes across the row
\item[\textit{``sustained''}] clearly visible in most or all frames
\end{promptitems}
A correct effect may legitimately be brief; this label is not a quality rating.

\promptreturn
\begin{lstlisting}[style=promptjson]
{"R_reading":"","O_reading":"",
 "per_property":{"appearance":"match|partial|mismatch","color":"match|partial|mismatch",
                 "dynamics":"match|partial|mismatch","integration":"match|partial|mismatch"},
 "localization":"correct|wrong|leaks",
 "effect_duration":"absent|brief|intermittent|sustained",
 "reasoning":""}
\end{lstlisting}

  \end{tcolorbox}
  \begin{tcolorbox}[breakable, enhanced, title={VFX -- Source Preservation (SP)},
    fonttitle=\bfseries\small, coltitle=white, colbacktitle=promptframe,
    colframe=promptframe, colback=promptback, boxrule=0.6pt, arc=1.5mm,
    left=5pt, right=5pt, top=4pt, bottom=4pt, before skip=8pt, after skip=10pt,
    fontupper=\fontsize{7.5}{9}\selectfont, parbox=false, label=prompt:vfx-sp]
  \setlength{\parskip}{2pt}\raggedright
  \promptrole{System}
You judge preservation for local reference-based effect transfer: whether an edited video kept everything it was supposed to keep.

The requested edit is:
\begin{promptitems}
\item[\textit{instruction}] ``\{instruction\}''
\item[\textit{transferred attribute}] ``\{transferred\_attribute\}''
\end{promptitems}

O should be S's video with only the transferred attribute changed. Two kinds of failure matter:
\begin{promptitems}
\item[\textit{(a)}] something of S's was lost or altered when it should have stayed;
\item[\textit{(b)}] something of R's, other than the transferred attribute, turned up in O.
\end{promptitems}

The transferred attribute appearing in O is correct, however strongly it resembles R's version. Changes that are a physical consequence of it~--- light it casts, reflections, shadows, what it occludes~--- are permitted and are not failures.

Judge from the pixels. Return one JSON object and nothing else. Do not output any numeric score.

\promptrole{User}
\promptattach{[attach: SOR\_rows] (one image, 3 rows x 6 columns, columns are time-aligned:\\ Top row    = S, the source, at 6 timestamps, earliest on the left\\ Middle row = O, the output, at the same 6 timestamps\\ Bottom row = R, the reference, at the same 6 timestamps\\ Read along a row to see how that video evolves over time.\\ Compare O with S by looking one row up, and O with R one row down.)}

\prompthead{Part A} one overall judgement for the whole clip. Compare the middle row against the top row. Mentally set aside everything the edit was allowed to change. Ignoring that completely, does what remains still look like the same video?
\begin{promptitems}
\item[\textit{``preservation\_overall''}] ``preserved'' \textbar{} ``partly\_changed'' \textbar{} ``different\_video''
\item[\textit{``overall\_reason''}] one sentence saying what drove that answer.
\end{promptitems}

\prompthead{Part B} attribute-by-attribute comparison.
\prompthead{Step B1} Read the top row (S) and list the key attributes S actually has. Use these slots, skipping any that do not apply to this clip:
\begin{promptitems}
\item[\textit{setting}] where it takes place: background, terrain, architecture, interior
\item[\textit{subjects}] the people / animals / main entities present, and their identity
\item[\textit{objects}] notable props and objects
\item[\textit{layout}] spatial arrangement, composition, framing
\item[\textit{camera}] camera position and how it moves along the row
\item[\textit{lighting}] lighting direction, tone, time of day
\item[\textit{action}] what happens across the six columns
\end{promptitems}
Describe S's version of each in a few concrete words $\rightarrow$ ``in\_source''

\prompthead{Step B2} For that same attribute, compare the middle row (O) against the top row (S) and against the bottom row (R), then choose one verdict:
\begin{promptitems}
\item[\textit{``preserved''}] O keeps S's version of it
\item[\textit{``side\_effect''}] changed only as an unavoidable consequence of the transferred attribute (light spill, reflection, shadow, occlusion)
\item[\textit{``altered''}] changed into something that is neither S's nor R's
\item[\textit{``from\_reference''}] O's version of it now matches R's instead of S's
\item[\textit{``absent''}] it is gone from O
\end{promptitems}
State your certainty~--- ``clear'' \textbar{} ``possible''~--- and describe what you actually see in O in a few concrete words $\rightarrow$ ``in\_output''

Do not list the transferred attribute itself as an attribute.

\promptreturn
\begin{lstlisting}[style=promptjson]
{"preservation_overall":"preserved|partly_changed|different_video",
 "overall_reason":"",
 "attributes":[{"name":"setting|subjects|objects|layout|camera|lighting|action",
                "in_source":"","in_output":"",
                "verdict":"preserved|side_effect|altered|from_reference|absent",
                "certainty":"clear|possible"}]}
\end{lstlisting}

  \end{tcolorbox}

\Needspace{20\baselineskip}
\subsection{Visual Quality}
\label{app:prompts-vq}
The VQ prompt is shared by all categories and receives only six frames of the output; withholding the source and reference prevents the judge from conflating visual quality with reference similarity.
  \begin{tcolorbox}[breakable, enhanced, title={All Categories -- Visual Quality (VQ)},
    fonttitle=\bfseries\small, coltitle=white, colbacktitle=promptframe,
    colframe=promptframe, colback=promptback, boxrule=0.6pt, arc=1.5mm,
    left=5pt, right=5pt, top=4pt, bottom=4pt, before skip=8pt, after skip=10pt,
    fontupper=\fontsize{7.5}{9}\selectfont, parbox=false, label=prompt:vq]
  \setlength{\parskip}{2pt}\raggedright
  \promptrole{System}
You assess the intrinsic quality of one video clip. Your job is to enumerate concrete, locatable defects~--- things you can actually point at in the frames. Do not give an overall impression, and do not rate the clip. Use the exact words offered below for ``property'' and ``obviousness''; do not substitute synonyms. Return one JSON object and nothing else. Do not output any numeric score.

\promptrole{User}
\promptattach{[attach: clip\_strip] (6 frames of the clip, earliest on the left)}

\prompthead{Step 1} Describe what the clip shows, in one or two sentences. $\rightarrow$ ``observation''

\prompthead{Step 2} Hunt for defects. Emit one entry per defect you can actually point at. Emit as many entries as you find, or none at all $\rightarrow$ ``defects''
\begin{promptitems}
\item[\textit{``property''}] exactly one of the three, the one it damages most:
\begin{promptitems}
\item[\textit{appeal}] aesthetic craft: composition, lighting, color, framing, overall production value
\item[\textit{plausibility}] does it look real: warping, melting, extra or missing limbs/fingers, broken text, flicker, ghosting, smeared texture, impossible anatomy or physics
\item[\textit{coherence}] does it hold together across the 6 frames: identity drift, objects appearing/vanishing, lighting or color jumps, scale or perspective inconsistency
\end{promptitems}
\item[\textit{``what''}] the defect and where it is, concretely (e.g. ``the left hand blurs into three fingers in frames 3-4'')
\item[\textit{``obviousness''}] exactly one of:
\begin{promptitems}
\item[\textit{``subtle''}] you had to look for it
\item[\textit{``noticeable''}] you would see it on a normal viewing
\item[\textit{``obvious''}] it dominates the clip; a viewer would reject it
\end{promptitems}
\end{promptitems}
If you genuinely find no defect for a property, emit nothing for it. An empty list is a valid and meaningful answer~--- do not invent defects to fill the list.

\prompthead{Step 3} ``defect\_summary'': one short phrase naming the worst defect you listed, or ``none'' if the list is empty.

\prompthead{Step 4} ``reasoning'': one or two sentences on why you assigned those obviousness levels.

\promptreturn
\begin{lstlisting}[style=promptjson]
{"observation":"",
 "defects":[{"property":"appeal|plausibility|coherence","what":"",
             "obviousness":"subtle|noticeable|obvious"}],
 "defect_summary":"",
 "reasoning":""}
\end{lstlisting}

  \end{tcolorbox}

\end{document}